\def\year{2021}\relax
\documentclass[letterpaper]{article} 
\usepackage{aaai21}  
\usepackage{times}  
\usepackage{helvet} 
\usepackage{courier}  
\usepackage[hyphens]{url}  
\usepackage{graphicx} 
\usepackage{natbib}  
\usepackage{caption} 
\usepackage{amsmath}
\usepackage{subfiles}
\usepackage[ruled,vlined]{algorithm2e}
\usepackage[framemethod=tikz]{mdframed}
\usepackage{tikz}
\usepackage{footmisc}
\usetikzlibrary{scopes, matrix, positioning}
\usepackage{booktabs}
\usepackage{amssymb}

\DeclareMathOperator*{\argmin}{arg\,min}

\newcommand{\setlabel}[1]{\edef\@currentlabel{#1}\label}

\title{Scaling Guarantees for Nearest Counterfactual Explanations}
\author{
    Kiarash Mohammadi\textsuperscript{\rm 1},
    Amir-Hossein Karimi\textsuperscript{\rm 2},
    Gilles Barthe\textsuperscript{\rm 3},
    Isabel Valera\textsuperscript{\rm 4}
    \\
}
\affiliations{
    \textsuperscript{\rm 1}MPI for Intelligent Systems and Ferdowsi University of Mashhad, \textsuperscript{\rm 2}MPI for Intelligent Systems and ETH-Z\"urich\\
    \textsuperscript{\rm 3}MPI for Security and Privacy and IMDEA Software Institute, \textsuperscript{\rm 4}MPI for Software Systems and Saarland University\\
    \{kiarash.mohammadi, amir\}@tue.mpg.de,
    gjbarthe@gmail.com,
    ivalera@cs.uni-saarland.de
}

\begin{document}

\maketitle

\begin{abstract}
   Counterfactual explanations (CFE) are being widely used to explain algorithmic decisions, especially in consequential decision-making contexts (e.g., loan approval or pretrial bail).
   In this context, CFEs aim to provide individuals affected by an algorithmic decision with the most similar individual (i.e., \emph{nearest} individual) with a different  outcome.
   However, while an increasing number of works propose algorithms to compute CFEs, 
   such approaches either lack in optimality of distance (i.e., they do not return the  \emph{nearest} individual) and perfect coverage (i.e., they do not provide a CFE for \emph{all} individuals); or they do not scale to complex models such as neural networks.
   In this work, we provide a framework based on Mixed-Integer Programming (MIP) to compute nearest counterfactual explanations  for the outcomes of neural networks, with both provable guarantees and  runtimes comparable 
   to gradient-based approaches.
   Our experiments on the Adult, COMPAS, and Credit datasets show that, in contrast with previous methods, our approach allows for efficiently computing diverse CFEs with both distance guarantees and perfect coverage.  
\end{abstract}

\section{Introduction}
\label{01_introduction}

Machine learning models are increasingly being used to assist in {semi}-automated prediction and decision-making for consequential scenarios such as pretrial bail and loan approval. Specifically, end-to-end trained models such as (deep) neural networks \cite{dl} (with non-linearities such as ReLU) have proven effective at learning and discovering complex non-linear patterns and relations in the data, and hence are becoming widely deployed. However, predictive power often comes at the cost of loss in interpretability \cite{rudin2018stop}, i.e., our ability to understand not only the decision made, but also the process by which the decision was deduced. Importantly, interpretability can assay the safe, robust, privacy-preserving, fair, and causally consistent nature of this decision-making \cite{doshi2017towards}.

Inspired by this, Counterfactual Explanations (CFEs) are introduced to provide individuals with an understanding of their situation in relation to a close hypothetical scenario in which they would have been treated favorably. As for the process of generating CFEs, a number of criteria are of concern: i) optimal distance, i.e., \emph{nearest} explanation; ii) perfect coverage, i.e., providing \emph{all} individuals with an explanation; iii) support for expressive models (e.g. neural networks); iv) efficient runtime; v) support for heterogeneous input spaces; and, vi) qualitative features such as actionability, plausibility, diversity, sparsity, etc.
While all these criteria have been discussed in previous works on CFE generation \cite{verma_survey, amir_survey}, existing approaches however lack in at least one of them. 

On one hand, providing the explanations with provable guarantees on the objectives (e.g., the proximity to the factual sample) has been studied by reducing the problem to a Satisfiability Modulo Theories (SMT) problem \cite{mace, mint} or to a Mixed-Integer Programming (MIP) problem \cite{efficient_search, dace, AR}. These approaches could theoretically be extended to support many classes of models, however, in practice this has only been demonstrated for simple classes of models, being high runtimes their main bottleneck. As an example, \citet{mace} show that even for reasonably small Neural Networks (NNs) (e.g. 20 neurons) the backend SMT solver might never terminate. In contrast, MIP-based approaches, however, so far ignore the class of NN models but instead work with simple linear~\cite{efficient_search, AR} or tree-based~\cite{dace} models, emphasizing  qualitative metrics of the explanations. 
On the other hand, counterfactual explanations can  be efficiently generated for (differentiable) NN models using gradient-based optimization techniques~\cite{dice}. However, while such approaches do work efficiently for NNs, they do not provide any guarantees in terms of distance or coverage. Moreover, they also suffer from limitations to incorporate qualitative aspects of CFE 
such as actionability constraints--e.g., an input feature capturing individuals' age is  only actionable in one direction, i.e., an individual can only increase her age.
Conclusively, previous approaches for CFE generation either ignore the class of neural models or cannot provide the aforementioned guarantees; the exception being MACE \cite{mace} which suffers from exponentially high runtimes. While NNs are becoming increasingly popular to adopt by stake-holders as a flexible non-linear model, an efficient approach with guarantees is necessary for explaining their decisions.

A similar problem to CFEs, in terms of formulation as a constrained optimization problem, is the generation of adversarial examples for NNs. This problem has been broadly addressed  by the NN verification community \cite{nnVerifSurvey}, where both SMT- and MIP-based approaches have been explored to efficiently solve the problem of finding adversarial examples in ReLU-activated NNs which is, in fact, shown to be NP-complete~\cite{reluplex}.
It is, however, important to note that while these two problems are formally similar and ideas can be exchanged among them, they are semantically and practically different \cite{wachter2017counterfactual}.  Thus, approaches to handle adversarial examples in NNs cannot be directly applied to generate CFEs \cite{cfe_adv_diff}.

In this work, we extend the ideas and tools from the NN verification community to develop an efficient framework to compute CFEs for ReLU-activated NN models, to provide distance and coverage guarantees, as well as to accommodate for previously discussed qualitative features. 
Specifically, we first propose three efficient approaches to search for a CFE within a given interval in the input feature space: whereas the first approach relies on SMT solvers as the backend, the other two approaches formulate the problem as a MIP and differ in the way that the CFE distance is optimized. 
All the three approaches make use of a linear approximation of the ReLU-NNs~\cite{planet} to compute bounds on the hidden units of the NN, given bounds on both the input feature space and/or distance. We then describe how to incorporate several qualitative features in our framework, including heterogeneous distance functions, as well as diversity and plausibility constraints~\cite{dace, efficient_search}.

Finally, we experiment our approaches on the before-mentioned criteria and compare against SMT- and gradient-based approaches that support NNs.
Table~\ref{tab:comparison} summarizes the fulfillment of different criteria in CFE generation by our approach in comparison with previous (SMT-, gradient-, and MIP-based) approaches. 
Our empirical results confirm a significant improvement in runtime efficiency, yielding novel MIP-based approaches for CFE generation on the class of NN models. Importantly, in addition to efficiently generating CFEs, our presented approaches are optimal in distance and perfect in coverage. This efficiency even allows for generating \emph{sets} of counterfactuals meeting different criteria, as we show by generating sets of diverse CFEs. Hence, while up to date, runtimes were the main bottleneck for CFE generation with guarantees for NN architectures, our MIP approach performs even faster than gradient-based optimization for NNs at the scale of consequential decision-making scenarios.

\begin{table*}[t]
\centering \fontsize{9}{11}\selectfont
  \begin{tabular}{lcccccc}
    \toprule
    Method & Opt. Distance & 100\% Coverage & Efficiency & Neural Models & Qualitative Features & Complex Constraints\\
    \midrule
    Our approach & \checkmark & \checkmark & \checkmark & \checkmark & \checkmark & \checkmark\\
    MACE \footnote{\cite{mace}} & \checkmark & \checkmark &  & \checkmark & \checkmark & \checkmark\\
    DiCE \footnote{\cite{dice}} &  &  & \checkmark & \checkmark & \checkmark & \\
    Efficient Search \footnote{\cite{efficient_search}} & \checkmark & \checkmark & \checkmark &  & \checkmark & \checkmark\\
  \bottomrule
\end{tabular}
\caption{Comparison of related work with our approach.\\
        \footnotesize{
        \textsuperscript{\rm 1} \citet{mace},
        \textsuperscript{\rm 2} \citet{dice},
        \textsuperscript{\rm 3} \citet{efficient_search}
        }}
\label{tab:comparison}
\end{table*}
\addtocounter{footnote}{-1}\let\thefootnote\svthefootnote
\addtocounter{footnote}{-1}\let\thefootnote\svthefootnote
\addtocounter{footnote}{-1}\let\thefootnote\svthefootnote

\section{Background}

We first introduce counterfactual explanations and two ways of formulating the problem, through optimization and verification. We then explain how the neural network model can be encoded within frameworks capable of solving the counterfactual explanation generation problem exactly and with guarantees.


\subsection{Counterfactual Explanations}
Assume that we are given a trained binary classifier $h: \mathcal{X} \rightarrow \rm I\!R$ that determines a positive outcome when $h(\bold{x}) \geq 0$ and a negative outcome when $h(\bold{x}) < 0$, deciding, e.g., whether an individual is eligible to receive a loan or not.
Consider an individual $\bold{x}^F$ where $h(\bold{x}^F) < 0$ (loan denial); for this individual, we would like to offer an answer to the question "What would have to be different for you to achieve a positive outcome next time?" \footnote{It is commonly assumed that the model is fixed and does not change over time.}
Answers to this question may be offered as a feature vector corresponding to an (hypothetical) individual on the other side of the decision boundary, and is referred to as a \emph{counterfactual explanation} (CFE).

There are a number of criteria/constraints that a CFE should satisfy to be useful for the individual~\cite{wachter2017counterfactual}.
A CFE should ideally be as similar as possible to the individual's current scenario (the factual instance), corresponding to the smallest change in the individual's situation that would favorably alter their prediction.
Furthermore, the change in features and the resulting counterfactual instance must satisfy additional \emph{feasibility} and \emph{plausibility} constraints, respectively.
For instance, a change in features that would require the individual to decrease their age would be \emph{infeasible (a.k.a. non-actionable)}.
Relatedly, we must make sure that the alternative scenario lies within the heterogeneous input space (i.e., is \emph{plausible}) since in the consequential decision-making domains, we typically work with mixed data types with a variety of statistical properties, such as age, race, bank balance, etc.

These requirements can be made more precise by assuming a notion of distance $dist$ between inputs, as well as predicates $Plausible$ and $Actionable$ for plausibility and actionability.

\subsubsection{\textbf{CFE Optimization Formulation}}
Counterfactual explanations can be modelled as a constrained optimization problem:
\begin{equation}
    \begin{split}
        \bold{x}^{CFE} \in \argmin_{\bold{x} \in \mathcal{X}} \qquad dist(\bold{x}, \bold{x}^F)\\
        s.t. \hspace{68pt} h(\bold{x}) \geq 0\\
        \bold{x} \in Plausible\\
        \bold{x} \in Actionable
    \end{split}
\end{equation}
The above optimization problem can be solved using Gradient Decent (GD) or linear programming, depending on the objective function and the constraints, and yields the closest input $\bold{x}^{CFE}$ (with respect to $\bold{x}^{F}$) that is plausible, actionable, and makes the decision of $h$ flip.

\subsubsection{\textbf{CFE Verification Formulation}}
Counterfactual explanations can be modelled as a satisfaction problem:
\begin{equation}
    \begin{split}
        \exists \bold{x}. dist(\bold{x}, \bold{x}^F)\leq\delta \\
        \hspace{68pt} h(\bold{x}) \geq 0\\
        \bold{x} \in Plausible\\
        \bold{x} \in Actionable
    \end{split}
\end{equation}
where $\delta$ is a distance threshold. The above satisfaction problem guarantees the existence of a counterfactual that is plausible, actionable, and within distance $\delta$ of $\bold{x}^F$. Using a suitable search strategy over $\delta$, it is then also possible to minimize $\delta$ (to an arbitrary precision) and find the \emph{nearest} counterfactual explanation. For example, MACE \cite{mace} encodes the above formulation using First-order logic and uses an SMT solver to find a series of counterfactuals within a binary search that minimizes $\delta$.

The precise formulation of the satisfaction problem depends on an encoding of $h$. Specifically, one must encode the classifier $h$ in the language of logic. While the encodings are theoretically well-understood, it is crucial to choose an encoding that guarantees the scalability of the method. Indeed, even for the simplest models, such as decision trees, naive encodings lead to verification tasks that exceed the capabilities of current tools. An important challenge is thus to develop efficient encodings of other models, and in particular of  NNs.
%


\subsection{Encoding NNs using SMT and MIP}

Outside of the domain of consequential decision-making, similar formulations to the CFE problem can be seen in the problem of \textit{adversarial examples}~\cite{papernot2017practical, moosavi2017universal, carlini2017towards}.
Here, there is a well-studied line of research towards verifying different properties of neural networks \cite{nnVerifSurvey}, such as robustness towards adversarial examples.
In this regard, many works focus on proving that a property holds or a counterexample exists. Among these works, many rely on SMT solvers, MIP-based optimization, or both \cite{planet, reluplex, unified}.

Neural network verification task (for ReLU-activated NNs) is shown to be NP-complete \cite{reluplex}. Different works, thus, try to make use of some properties and guide the search process in a way to work better than conventional off-the-shelf solvers or optimizers.
Subsequently, we try to do the same for CFE generation and extend the previous work, MACE \cite{mace}, to work better than using off-the-shelf solvers in a straight-forward manner.
This happens through, e.g., guiding the search process by gradually increasing the distance within which we are looking for a counterfactual explanation, keeping the distance interval as small as possible to prune domains efficiently.

In the following, we explain how to represent NNs using First-order predicate logic formulae and as an MIP that provide bounds on the optimization variables, later resulting in efficient domain pruning within the search for CFEs.
%


\subsubsection{\textbf{First-order Logic (SMT) Encoding of Neural Networks}}
\label{smt_encoding}
It is rather straight-forward to encode neural networks using a First-order logic representation that is acceptable by Satisfiability Modulo Theories  (SMT) oracles \cite{mace}. Figure \ref{fig:smt_encoding} shows this through an example ($\hat{z}_1$ and $\hat{z}_2$ represent the post-ReLU values).

\tikzset{
  mymx/.style={matrix of math nodes,nodes=myball,column sep=2.em,row sep=-1ex},
  myball/.style={draw,circle,inner sep=4pt},
  mylabel/.style={near start,sloped,fill=white,inner sep=1pt,outer sep=1pt,below,
    execute at begin node={$\scriptstyle},execute at end node={$}},
  plain/.style={draw=none,fill=none},
  sel/.append style={fill=green!10},
  prevsel/.append style={fill=red!10},
  route/.style={-latex,thick},
  selroute/.style={route,blue!50!green}
}

\begin{figure}[h]
\vspace{-20pt}
\begin{minipage}[b]{0.28\linewidth}
\begin{center}
  \begin{tikzpicture}
  \hspace{-10pt}
    \matrix[mymx] (mx) {
      |[plain]| x_1 \\
                    & z_1    \\
      |[plain]| x_2 &         & z_3  \\
                    & z_2     \\
      |[plain]| x_3 \\
    };
    {[route]
      \draw (mx-1-1) -- (mx-2-2) node[mylabel, above] { 1 };
      \draw (mx-1-1) -- (mx-4-2) node[mylabel, above] { 2 };
      \draw (mx-3-1) -- (mx-2-2) node[mylabel, above] { -1 };
      \draw (mx-3-1) -- (mx-4-2) node[mylabel, above] { 0 };
      \draw (mx-5-1) -- (mx-2-2) node[mylabel, above] { 0 };
      \draw (mx-5-1) -- (mx-4-2) node[mylabel, above] { -1 };
      \draw (mx-2-2) -- (mx-3-3) node[mylabel, midway, above] { \hat{z}_1, \,-1 };
      \draw (mx-4-2) -- (mx-3-3) node[mylabel, midway, below] { \hat{z}_2, \,1 };
    }
  \end{tikzpicture}
\end{center}
\end{minipage}
\begin{minipage}[b]{0.71\linewidth}
\begin{align*}
    \phi_{f}(& x)=\\[-1ex]
    (z_1 = x_1 - x_2) & \wedge (z_2 = 2x_1 - x_3)\\[-1ex]
    \wedge ((\hat{z}_1 = z_1 \wedge z_1 \geq 0 &) \vee (\hat{z}_1 = 0 \wedge z_1 < 0))\\[-1ex]
    \wedge ((\hat{z}_2 = z_2 \wedge z_2 \geq 0 &) \vee (\hat{z}_2 = 0 \wedge z_2 < 0))\\[-1ex]
    \wedge (z_3 = &- \hat{z}_1 + \hat{z}_2) 
\end{align*}
\end{minipage}
\caption{A ReLU-activated neural network and its corresponding logic formula}
\label{fig:smt_encoding}
\end{figure}
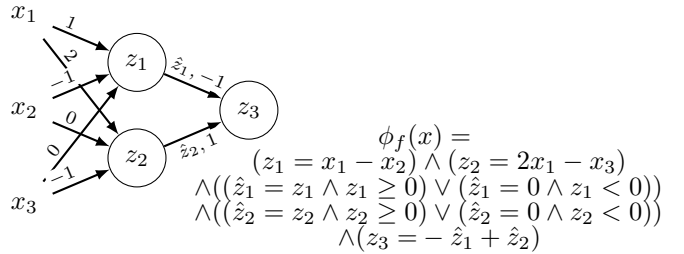


\subsubsection{\textbf{Unbounded Mixed-integer Program Encoding of Neural Networks}}
\label{mip_encoding}
We try to be faithful to the notation from \citet{nnVerifSurvey}. Consider an $n$-layer single-output feed-forward neural network (NN) with ReLU activations after each hidden layer that represents the function $h(\bold{x})$. The width of each layer is $k_i$ and $\bold{z}_i$ is the vector of dimension $k_i$ which represents layer $i$ where $i \in \{1, 2, ..., n\}$. While $\bold{z}_i$ represents the pre-ReLU activations, $\hat{\bold{z}}_i$ is the values after ReLUs have been applied. Finally, $\boldsymbol{\delta}_i$ are vectors of binary variables indicating the state of each ReLU; 0 for inactive and 1 for activated ReLUs.

There are multiple ways to encode neural networks as MIPs in the NN verification literature, each proposing different encodings for ReLU activations. A generic form is as follows. For $ i \in \{1,..., n\}$ and $ j \in \{1,..., k_i\}$:
\begin{subequations} \label{eq:mip_unbound}
    \begin{equation} \label{eq:mip_unbound_a}
        \bold{z}_{i} = \bold{W}_{i}\hat{\bold{z}}_{i-1} + \bold{b}_{i}
    \end{equation}
    \begin{equation} \label{eq:mip_unbound_b}
    \begin{split}
         \boldsymbol{\delta}_i \in \{0, 1\}^{k_i},~ \hat{\bold{z}}_i = \bold{z}_i \cdot \boldsymbol{\delta}_i,\\
         \delta_{i, j} = 1 \Rightarrow z_{i, j} \geq 0,\\
         \delta_{i, j} = 0 \Rightarrow z_{i, j} < 0
    \end{split}
    \end{equation}
\end{subequations}

\noindent The first part \eqref{eq:mip_unbound_a} is simply the linear affine of weights and the second part \eqref{eq:mip_unbound_b} encodes the following ReLUs using the introduced binary variables for each ReLU. We refer to this as the \textit{unbounded} MIP encoding.


\subsubsection{\textbf{Bounded Mixed-integer Program Encoding of Neural Networks}}

\citet{unified} suggest that most NN verifiers, based on either SMT or MIP solvers, are indeed a variation of Branch-and-Bound (B\&B) optimization. This understanding implies that limiting the bounds of the variables of the optimization problem is a very effective heuristic. Moreover, the extra constraints of the CFE generation problem -- making the verification formulation difficult to solve -- might actually help tightening the bounds, and thus, result in an effective pruning of the domains of the optimization problem. 
We will thus, change the generic ReLU formulation \eqref{eq:mip_unbound_b} and adopt the bounded encoding proposed by \citet{mip_nn}, i.e., for $ i \in \{1,..., n\}$:

\begin{subequations} \label{eq:mip}
    \begin{equation} \label{eq:mip_a}
        \bold{z}_{i} = \bold{W}_{i}\hat{\bold{z}}_{i-1} + \bold{b}_{i}
    \end{equation}
    \begin{equation} \label{eq:mip_b}
    \begin{split}
        \boldsymbol{\delta}_i \in \{0, 1\}^{k_i}, \quad \hat{\bold{z}}_i \geqslant 0, \quad \hat{\bold{z}}_i \leqslant \bold{u}_i \cdot \boldsymbol{\delta}_i,\\
        \hat{\bold{z}}_i \geqslant \bold{z}_i, \quad \hat{\bold{z}}_i \leqslant \bold{z}_i - \bold{l}_i \cdot (1 - \boldsymbol{\delta}_i)
    \end{split}
    \end{equation}
\end{subequations}

Note that the linear part \eqref{eq:mip_a} is the same as \eqref{eq:mip_unbound_a} and also  note that this is still an exact encoding of NNs using MIP since $\delta_{i,j} = 0 \Leftrightarrow	\hat{z}_{i,j} = 0$ and $\delta_{i,j} = 1 \Leftrightarrow \hat{z}_{i,j} = z_{i,j}$.
This encoding relies on $\bold{l}_i$ and $\bold{u}_i$, vectors indicating the lower and upper bounds of the values of the hidden units at layer $i$. We remind that tight bounds can be very effective in domain pruning when solving the mixed-integer program. Here, we introduce two ways to obtain such bounds and complete the MIP formulation \eqref{eq:mip} for CFEs: first, using interval arithmetic \cite{interval_arith}, and second, using an approximation of ReLUs that results in tighter bounds. In both cases, we assume that we have initial lower/upper bounds on the values of the input layer (e.g., derived from the dataset). This is a valid assumption since real-world features such as age or income do have bounds.
%


\subsubsection{\textbf{Interval arithmetic.}} 
By using interval arithmetic \cite{interval_arith}, having the bounds at layer $i-1$, we can compute the bounds for the $j$-th neuron from the $i$-th layer ($z_{i, j}$) as:
\begin{equation} \label{eq:interval_arith}
    \begin{aligned}
        l_{i, j} = \Sigma_{t=1}^{k_{i-1}} & ( max(W_{i, j, t}, 0) \cdot l_{i-1, t}\\ 
        & + min(W_{i, j, t}, 0) \cdot u_{i-1, t}) + b_{i, j}\\
        u_{i, j} = \Sigma_{t=1}^{k_{i-1}} & ( max(W_{i, j, t}, 0) \cdot u_{i-1, t}\\ 
        & + min(W_{i, j, t}, 0) \cdot l_{i-1, t}) + b_{i, j}
    \end{aligned}
\end{equation}

\noindent The post-ReLU bounds (for $\hat{z}_{i, j}$) are obtained simply by applying a ReLU on these bounds.

This is applied layer-by-layer and the bounds for all hidden units are computed recursively starting from the input layer. Unfortunately, although better than having no bounds at all, these bounds quickly become loose as we go deeper in the network. The reason is that in each layer $i$, each neuron is choosing a worst-case bound (lower or upper) from the neurons of the previous layer $i-1$, independently from the rest of the neurons in layer $i$, causing conflicts in the choice of the lower or upper bound for some neurons in layer $i-1$.\footnote{Refer to the Appendix for more explanation by an example. \label{fn:example}}


\subsubsection{\textbf{Linear over-approximation of ReLUs.}}
\label{linearized_net_approximator}
To compute tighter bounds than interval arithmetic, we first adopt the linear \emph{over-approximation} of ReLUs proposed in~\cite{planet} to replace \eqref{eq:mip_unbound_b}, i.e., for $ i \in \{1,..., n\}$ and $ j \in \{1,..., k_i\}$:
\begin{subequations} \label{eq:approx}
    \begin{equation} \label{eq:approx_a}
        \bold{z}_{i} = \bold{W}_{i}\hat{\bold{z}}_{i-1} + \bold{b}_{i}
    \end{equation}
    \vspace{-16pt}
    \begin{equation} \label{eq:approx_b}
        \hat{\bold{z}}_i \geqslant \bold{z}_i, \quad \hat{\bold{z}}_i \geqslant 0, \quad \hat{z}_{i,j} \leqslant u_{i,j} \frac{z_{i,j} - l_{i,j}}{u_{i,j} - l_{i,j}} 
    \end{equation}
\end{subequations}

\noindent Again, the linear part \eqref{eq:approx_a} is the same as \eqref{eq:mip_unbound_a}. For the ReLU part \eqref{eq:mip_unbound_b}, the binary variables encoding the ReLUs in an exact way are removed and, instead, a linear over-approximation term has been replaced \eqref{eq:approx_b}. This results in a fully linear MIP system without the ReLU binary variables, whose optimization for different objectives can be performed efficiently.

As before, the bounds are recursively computed in a layer-by-layer manner, and the constraints of the linearized network \eqref{eq:approx} are added to the MIP system progressively. At each layer $i$, first, \eqref{eq:approx_a} is added with bounds of the variables computed using simple interval arithmetic from the tight bounds computed for the previous layer. Then, to find better bounds than simple interval arithmetic, having included all the constraints up until this layer, two MIPs are solved for each hidden unit: one with the objective of maximizing the value of the unit to compute an upper bound, and a similar one for computing the lower bound. Finally, the ReLU constraints \eqref{eq:approx_b} for this layer are added with the just-computed tight bounds.\footref{fn:example}
Note that while we have opted for the ReLU activation function as a common source of non-linearity, any activation function that can be approximated by piece-wise linear functions is applicable, e.g., Max-Pooling \cite{planet}.

We build upon an implementation from \citet{unified} for this purpose. Obtaining tight bounds here relies on how small the domains of the input variables are; keeping the input domains small enough will result in tighter bounds for other variables. This will be discussed in more detail in the next section.

\section{CFE Generation}

In this section, we propose three approaches towards CFE generation for neural networks. All the approaches rely on the linearized network approximations described in the previous section, which provide tight lower and upper bounds on the values of the hidden units.
Below, we first explain the search strategy on the distance of the nearest CFE and the way lower/upper bounds on the input and hidden units are computed within this search.
Then, we introduce three approaches towards efficient nearest CFE generation for neural networks.


\subsection{Preliminaries}


\subsubsection{\textbf{Exponential Search Strategy.}}
In order to optimize the distance towards finding the nearest CFE, we implement an exponential search strategy \cite{exp_search}. 
W.l.o.g., we assume here that the input space is normalized and lies within the [0, 1] interval. 
Because the interval of the input layer determines those of later layers, we initiate our search with a small distance interval, whose lower and upper bound are set respectively to 0 and an (arbitrarily) small $\epsilon$.
We then exponentially increase the search interval until a CFE is found. 
Finally, a simple binary search is performed on the interval where the CFE was found to look for the nearest CFE. 
The overall scheme for the exponential search is summarized in Algorithm \ref{alg:exp_search}.

\begin{algorithm}
\SetAlgoLined
\KwIn{$\bold{N}$, $\bold{x}^F$, $\epsilon$}
\KwOut{closest\_CFE}
\SetKwFunction{findCFE}{findCFE}
\SetKwFunction{binarySearch}{binarySearch}
 $[lb_{dist}, ub_{dist}] \gets [0, \epsilon]$\;
 \While{$\findCFE(\bold{N}, \bold{x}^F, lb_{dist}, ub_{dist})$ is None}{
  $lb_{dist} \gets ub_{dist}$\;
  $ub_{dist} \gets ub_{dist} \times 2$\;
 }
 closest\_CFE $\gets \binarySearch(\bold{N}, \bold{x}^F, \epsilon, lb_{dist}, ub_{dist})$\;
 \Return closest\_CFE\;
 \caption{Exponential Search Strategy}
 \label{alg:exp_search}
\end{algorithm}

Next, we  discuss how to compute bounds on both the input and hidden units of the network, which are necessary to efficiently implement the CFE search function, \texttt{findCFE} in Algorithm~\ref{alg:exp_search}.
%


\subsubsection{\textbf{Computing Bounds for Input and Hidden Units.}}
\label{all_boound_computation}
We leverage the network approximator based upon equation \eqref{eq:approx} to compute the bounds of the network input and hidden units for a given distance interval $[lb_{dist}, ub_{dist}]$. 
To this end, we first obtain the MIP encoding of the distance. 
Then, we optimize the MIP-encoded distance for each input variable, maximizing/minimizing each variable to obtain the lower/upper bounds of the input layer for the given distance interval.
Then, the input bounds are propagated in the NN to compute the bounds of hidden units. 
We include the distance constraints in the initial constraint set of the linearized network to help finding tighter bounds for the hidden units. Algorithm \ref{alg:lin_approximator} shows the overall scheme for this.

\begin{algorithm}
\SetAlgoLined
\KwIn{$\bold{N}$, $\bold{x}^F$, $lb_{dist}$, $ub_{dist}$}
\KwOut{$\bold{LB}_{net}$, $\bold{UB}_{net}$}
\SetKwFunction{getDistanceConstraints}{getDistanceConstraints}
\SetKwFunction{optimizeInputVars}{optimizeInputVars}
\SetKwFunction{linearizedNetApproximator}{linearizedNetApproximator}
 $\phi_{dist} \gets \getDistanceConstraints(\bold{N}, \bold{x}^F, lb_{dist}, ub_{dist})$\;
 $\bold{lb}_{inp}, \bold{ub}_{inp} \gets \optimizeInputVars(\bold{N}, \phi_{dist})$\;
 $\bold{LB}_{net}, \bold{UB}_{net} \gets \linearizedNetApproximator(\bold{N}, \bold{lb}_{inp}, \bold{ub}_{inp}, \phi_{dist})$\;
 \Return $\bold{LB}_{net}, \bold{UB}_{net}$\;
 \caption{Bounds Computation}
 \label{alg:lin_approximator}
\end{algorithm}


\subsection{Approaches}

In this section, we propose three efficient approaches to implement the CFE search function, \texttt{findCFE} in Algorithm~\ref{alg:exp_search}, for neural networks. 
The first approach relies on SMT solvers as backend and uses the bounds computation as a heuristic within each iteration of the exponential search (Algorithm \ref{alg:exp_search}). 
The second and third approaches instead rely on MIP solving to search for CFEs. The difference between them lies on the optimization of the distance -- while the second approach minimizes the CFE distance using the exponential search described above, the third approach includes the distance as objective within the MIP optimization framework. Next, we provide further details on the three approaches.

\subsubsection{\textbf{ReLU Elimination (MIP-SAT).}}
In this approach, we build upon MACE \cite{mace} (SMT solving in the backend) and use the bounds computation as a heuristic. Within each iteration of the exponential search (Algorithm \ref{alg:exp_search}), and given the distance interval, the bounds on the input and hidden units are computed using Algorithm \ref{alg:lin_approximator} and ReLUs with a fixed state are determined. A ReLU has a fixed state iff the value of the neuron before applying ReLU has either a lower bound greater than or equal to zero, or an upper bound less than or equal to zero.

The neural network, distance functions, as well as additional  constraints are primarily encoded as SMT formulae. 
For the NN bound computation,  the NN and distance constraints are encoded as MIPs, as described before.
Next, the ReLUs with a fixed-state  are removed from the initial SMT formula representing the NN. This means that, for an always-active ReLU, we will have $\hat{z}_i = z_i$ and for an always-inactive ReLU we will have $\hat{z}_i = 0$, instead of the initial ReLU clause: $(\hat{z}_i = z_i \wedge z_i \geq 0) \vee (\hat{z}_i = 0 \wedge z_i < 0)$. This is, basically, removing the disjunction associated to the ReLU states by fixing its value, saving the SMT solver the effort to branch over its cases. Finally, the SMT solver (Z3 solver \cite{z3} in our case) is called with the new formula to verify the existence of a CFE within the given distance interval.

Note that the ReLU clauses in the SMT representation of the neural network are exponentially expensive to handle for the SMT solver since it needs to branch over the cases. Thus, removing a subset of the RELU activations will reduce the run-time exponentially (as empirically shown in the experiments). Algorithm \ref{alg:mip_sat} shows the overall scheme for the proposed mixed MIP-SAT approach.

\begin{algorithm}
\SetAlgoLined
\KwIn{$\bold{N}$, $\bold{x}^F$, $lb_{dist}, ub_{dist}$}
\KwOut{CFE or None}
\SetKwFunction{getDistanceFormula}{getDistanceFormula}
\SetKwFunction{getPlausibilityFormula}{getPlausibilityFormula}
\SetKwFunction{getModelFormula}{getModelFormula}
\SetKwFunction{eliminateRelus}{eliminateRelus}
\SetKwFunction{computeBounds}{computeBounds}
\SetKwFunction{SAT}{SAT}
\SetKwFunction{findCFE}{findCFE}
 $\phi_{dist} \gets \getDistanceFormula(\bold{N}, \bold{x}^F, lb_{dist}, ub_{dist})$\;
 $\phi_{pls} \gets \getPlausibilityFormula(\bold{N})$\;
 $\phi_{N} \gets \getModelFormula(\bold{N})$\;
 $\bold{LB}_{net}, \bold{UB}_{net} \gets \computeBounds(\bold{N}, \bold{x}^F, lb_{dist}, ub_{dist})$\;
 $\phi_{N} \gets \eliminateRelus(\phi_{N}, \bold{LB}_{net}, \bold{UB}_{net})$\;
 \uIf{$\SAT(\phi_{N} \wedge \phi_{dist} \wedge \phi_{pls})$}
 {
    \Return CFE\;
 }
 \uElse{
    \Return None\;
 }
 \caption{The MIP-SAT approach -- \texttt{findCFE} in Algorithm \ref{alg:exp_search}}
 \label{alg:mip_sat}
\end{algorithm}


\subsubsection{\textbf{Output Optimization (MIP-EXP).}}
In this approach, we purely use a MIP-based optimization process (no SMT oracle), for which we deploy an optimization engine (Gurobi \cite{gurobi} in this case), building upon an implementation of \eqref{eq:mip} from~\citet{unified}.

As before, we assume that we are within an iteration of the exponential search (Algorithm \ref{alg:exp_search}) with a fixed distance interval $[lb_{dist}, ub_{dist}]$. 
First, Algorithm \ref{alg:lin_approximator} is called to compute tight lower/upper bounds for the input and hidden units of the network. Next, these bounds are used to obtain MIP encoding of the neural network as in \eqref{eq:mip}. 
Then the distance, as well as any other additional constraints (all explained in the next section), are added to MIP formulation. %
Finally, depending on the (predicted) label of the factual sample $\bold{x}^F$, the single output of the network is optimized. 
For instance, for a factual sample with a positive label, the output of the network will be minimized with a callback that interrupts the optimization as soon as a counterfactual with a negative output value is found. Otherwise, the lower bound of the output of the network for this factual sample and distance interval is greater than zero and no counterfactual exists. 
The overall scheme of the proposed MIP-EXP approach is shown in Algorithm \ref{alg:mip_exp}.

Note that this approach no longer uses an SMT oracle, but instead  relies on an optimization engine to solve a mixed-integer program with the single output of the network as its objective function. Thus, it can naturally be extended to multi-class classification by introducing a new variable in the MIP that preserves the maximum logit among class outputs on which the optimization objective is defined.

\begin{algorithm}
\SetAlgoLined
\KwIn{$\bold{N}$, $\bold{x}^F$, $lb_{dist}, ub_{dist}$}
\KwOut{CFE or None}
\SetKwFunction{getDistanceConstraints}{getDistanceConstraints}
\SetKwFunction{getPlausibilityConstraints}{getPlausibilityConstraints}
\SetKwFunction{computeBounds}{computeBounds}
\SetKwFunction{getModelConstraints}{getModelConstraints}
\SetKwFunction{optimize}{optimize}
\SetKwFunction{findCFE}{findCFE}
 $\phi_{dist} \gets \getDistanceConstraints(\bold{N}, \bold{x}^F, lb_{dist}, ub_{dist})$\;
 $\phi_{pls} \gets \getPlausibilityConstraints(\bold{N})$\;
 $\bold{LB}_{net}, \bold{UB}_{net} \gets \computeBounds(\bold{N}, \bold{x}^F, lb_{dist}, ub_{dist})$\;
 $\phi_{N} \gets \getModelConstraints(\bold{N}, \bold{LB}_{net}, \bold{UB}_{net})$ \tcp*{MIP encoding \ref{eq:mip}}
 \uIf{$\optimize(\phi_{N}, \phi_{dist}, \phi_{pls}, \bold{x}^F)$}
 {
    \Return CFE\;
 }
 \uElse{
    \Return None\;
 }
 \caption{The MIP-EXP approach -- \texttt{findCFE} in Algorithm \ref{alg:exp_search}}
 \label{alg:mip_exp}
\end{algorithm}


\subsubsection{\textbf{Distance Optimization (MIP-OBJ).}}
This is similar to the MIP-EXP approach except that we remove the outer loop (the exponential search of Algorithm \ref{alg:exp_search}) and the distance function is introduced as the objective function of the MIP to be minimized.

In this approach, which we refer to as MIP-OBJ, Algorithm \ref{alg:lin_approximator} is called to compute the bounds with the distance interval being $[0, 1]$. The computed bounds are placed within MIP encoding \eqref{eq:mip}. Since now the objective of the MIP is the distance function, we need to add a constraint as the \emph{counterfactual constraint} determining the single output of the network being negative or positive based on the (predicted) label of the factual sample. 
The whole problem is optimized (with an optimality gap of $\epsilon$ for the distance objective to be analogous to the other approaches) and the nearest CFE is found. Algorithm \ref{alg:mip_obj} shows the overall scheme of the  MIP-OBJ approach.

\begin{algorithm}
\SetAlgoLined
\KwIn{$\bold{N}$, $\bold{x}^F$, $lb_{dist}, ub_{dist}$}
\KwOut{CFE or None}
\SetKwFunction{getDistanceConstraints}{getDistanceConstraints}
\SetKwFunction{getPlausibilityConstraints}{getPlausibilityConstraints}
\SetKwFunction{getCounterfactualConstraint}{getCounterfactualConstraint}
\SetKwFunction{computeBounds}{computeBounds}
\SetKwFunction{getModelConstraints}{getModelConstraints}
\SetKwFunction{optimize}{optimize}
\SetKwFunction{findCFE}{findCFE}
 $obj \gets \getDistanceConstraints(\bold{N}, \bold{x}^F)$\;
 $\phi_{pls} \gets \getPlausibilityConstraints(\bold{N})$\;
 $\phi_{CFE} \gets \getCounterfactualConstraint(\bold{N}, \bold{x}^F)$\;
 $\bold{LB}_{net}, \bold{UB}_{net} \gets \computeBounds(\bold{N}, \bold{x}^F, 0, 1)$\ \tcp*{No distance limit}
 $\phi_{N} \gets \getModelConstraints(\bold{N}, \bold{LB}_{net}, \bold{UB}_{net})$ \tcp*{MIP encoding \ref{eq:mip}}
 $CFE \gets \optimize(\phi_{N}, \phi_{pls}, \phi_{CFE}, obj, \bold{x}^F)$\;
 \Return CFE\;
 \caption{The MIP-OBJ approach}
 \label{alg:mip_obj}
\end{algorithm}

\section{Distance Functions and Qualitative Features}

In this section ,we describe how the distance metric, as well qualitative features--such as plausibility, sparsity and diversity--can be encoded within the MIP framework. 
First, we provide details on the encoding of distance functions suitable for heterogeneous input features.
Second, in the context of plausibility,  we describe how to handle heterogeneous input spaces, i.e., input features with mixed data types.
Finally, we focus on a broadly studied qualitative property of CFEs, diversity. We would like to emphasize that previous MIP-based approaches have recognized the flexibility of mixed-integer programming in regards to encode a wide range of complex constraints and different qualitative features \cite{efficient_search, dace}, however, this  cannot be directly leveraged for NN models. We defer to future work to address a wider range of qualitative features for NN class of models.
%


\subsection{Distance Functions}
\label{sec:distance_functions}

In this section, we provide more details on the MIP encoding of heterogeneous distance functions.\footnote{For conciseness, the intermediate variables used to practically encode the functions within the MIP model are excluded here.} We provide details on an $\ell_1$ distance function (analogous to previous works \cite{wachter2017counterfactual}) while zero-, two-, and infinity-norms are supported in an analogous manner, each providing a different practical intuition for the proximity of the CFEs, e.g., $\ell_0$ used for sparsity. As described before, the distances are all range normalized and within the $[0, 1]$ interval.\\

\noindent \textbf{Integer-valued and real-valued features.} For an input vector $\bold{x}$ and factual sample $\bold{x}^F$ with such a feature at the $i$-th dimension, the normalized $\ell_1$ distance is computed in a straight-forward manner:

\begin{equation}
\label{eq:dist_1}
    dist_{real}(x_i, x^{F}_i) = \frac{|x_i - x^{F}_{i}|}{ub_i - lb_i}
\end{equation}

\noindent where $lb_i, ub_i$ are the scalar lower/upper bounds for $x_i$.\\

\noindent \textbf{Ordinal features}. For an input vector $\bold{x}$ and factual sample $\bold{x}^F$ with an ordinal feature $\bold{x}_i$ having $k$ levels, the normalized $\ell_1$ distance is computed in the following manner:

\begin{equation}
\label{eq:dist_2}
    dist_{ord}(\bold{x}_i, \bold{x}^{F}_i) = \frac{|\sum_{j=1}^{k} x_{i, j} - \sum_{j=1}^{k} x^F_{i, j}|}{k}
\end{equation}\\
\noindent \textbf{Categorical features}. For an input vector $\bold{x}$ and factual sample $\bold{x}^F$ with a categorical feature $\bold{x}_i$ having $k$ categories, the normalized $\ell_1$ distance is computed in the following manner:

\begin{equation}
\label{eq:dist_3}
    dist_{cat}(\bold{x}_i, \bold{x}^{F}_i) = \max_{1 \leq j \leq k} (x_{i, j} - x^{F}_{i,j})
\end{equation}\\
In the end, the total normalized $\ell_1$ distance between input vector $\bold{x}$ and factual sample $\bold{x}^F$ would be the normalized sum over distances of different data types \eqref{eq:dist_1}, \eqref{eq:dist_2}, \eqref{eq:dist_3}, $n_{real}, n_{ord}, n_{cat}$ being the number of features in each of the three groups above:

\begin{equation}
\label{eq:dist_4}
    \begin{split}
    dist(\bold{x}, \bold{x^{F}}) = \frac{1}{n_{real} + n_{ord} + n_{cat}}
    (\sum\limits_{i=1}^{n_{real}} dist_{real}(x_i, x^{F}_i) \\
    + \sum\limits_{i=1}^{n_{ord}} dist_{ord}(\bold{x}_i, \bold{x}^{F}_i) 
    + \sum\limits_{i=1}^{n_{cat}} dist_{cat}(\bold{x}_i, \bold{x}^{F}_i))
    \end{split}
\end{equation}\\
\noindent \textbf{Sparsity}. Sparsity can be interpreted as the $\ell_0$ distance function. It is encoded by introducing a number of intermediate binary variables each retaining whether or not a feature has changed its value and then summed over and normalized analogous to the described $\ell_1$ distance.
%


\subsection{Plausibility Constraints} \label{sec:plausibility_constraints}
In this section we explain plausibility constraints that guarantee the CFE lying within the same heterogeneous space as input.
Plausibility constraints for integer-valued, real-valued, and binary variables are naturally preserved by defining the right kind of variables within the MIP (or SMT) model.\\

\noindent \textbf{Ordinal features.} To guarantee that the CFEs are plausible in terms of ordinality of the ordinal features, for each such feature $\bold{f}$ with $k$ levels, we define $k$ binary variables $f_1, ..., f_k \in \{0, 1\}$ in the MIP model. For each set of these variables, the following constraints are added to the MIP model:

\begin{equation}
    f_1 \geq f_2, f_2 \geq f_3, ..., f_{k-1} \geq f_k
\end{equation}

\noindent This will guarantee that: $\not\exists ~ i ~ \text{s.t.} ~ f_{i+1} > f_i$.\\

\noindent \textbf{Categorical features.} We want to guarantee that in the produced CFE, for each categorical feature, only one category is chosen. For a categorical feature $\bold{f}$ with $k$ categories, we define $k$ binary variables $f_1, \dots, f_k \in \{0, 1\}$ in the MIP model. For each set of these variables, the following constraint is added to the MIP model:
\begin{equation}
    f_1 + f_2 + \dots + f_k = 1
\end{equation}
Since $f_i$'s are binary variables, this will guarantee that only one of them is 1 and others are 0, meaning that at most one category is active as desired.
%


\subsection{Diversity Constraints}
\label{sec:diversity}

Providing individuals with different, preferably diverse, counterfactuals can be beneficial in terms of providing alternative ways for the individuals to improve their outcome. Having different diverse (and close) counterfactuals, the individuals may find the most suitable way to achieve the preferred outcome while considering their own personal constraints, about which the explanation-provider might not be aware of.

As with other qualitative features, there are different ways for encoding diversity in the literature of CFE generation. Within the MIP-based approaches, \citet{efficient_search} encodes diversity simply as the newly generated CFE not being equal to the previously generated ones. Based on the evaluation criteria, this could fail to generate diverse CFEs, for example when the evaluation criteria is the mean of the pairwise distances of the ($k$) generated CFEs as DiCE \cite{dice} suggests. Among the gradient-based approaches, DiCE \cite{dice} accounts for diversity using determinantal point processes, i.e.,  it includes the determinant of the kernel matrix given the counterfactuals in the objective.

It is important to also take into account the distance of the generated set of diverse counterfactuals since it is necessary for this set to also be close to the individual for which it is being generated. Thus, it can be seen that there is an inherent tradeoff between diversity and distance. To account for this, we encode diversity as a set of constraints for each newly generated counterfactual to have a distance above a fixed threshold from each of the previously generated counterfactuals, while minimizing the distance to the factual sample. More specifically, the following set of constraints will be added before the search for the $i$-th CFE:

\begin{equation}
\label{eq:diversity}
    \begin{split}
        dist(\bold{x}_{1}^{CFE}, \bold{x}_{i}^{CFE}) \geq \delta\\
        \ldots\\
        dist(\bold{x}_{i-1}^{CFE}, \bold{x}_{i}^{CFE}) \geq \delta
    \end{split}
\end{equation}

Note that solving the MIP becomes progressively more expensive for each new counterfactual. We have implemented a version of our approach called MIP-DIVERSE for generating diverse counterfactuals using the above formulation.

\section{Experiments}
\label{05_experiments}

We conduct a number of quantitative and qualitative experiments to demonstrate our frameworks abilities relative to existing approaches: MACE~\cite{mace} \footnote{We use an improved version of MACE obtained from the official GitHub repository.} and DiCE~\cite{dice}.\footnote{We use default hyperparameters for DiCE, as obtained from the official GitHub repository of DiCE (commit @92530c7). In all but the diversity experiments that will follow, we set the diversity weight to zero since we are searching for only one CFE and want the focus only on proximity and flipping of the output. \label{fn:dice}}~
Following the motivation explained in the Introduction, we generate counterfactual explanations for fixed-width ReLU-activated fully-connected NN models of various sizes, having $N \times W + (D-1) \cdot W^2 + (D+1) \times W$ total parameters, $N$ being the input size, $W$ width, and $D$ depth.
To support consequential decision-making settings, we employ three widely used real-world datasets from the counterfactual explanations literature: Adult ($d=51$) \cite{adult_dataset}, COMPAS ($d=7$) \cite{propublica_compas}, and Credit ($d=20$) \cite{bache2013uci}.
Finally, all approaches are evaluated and compared on their optimality of distance, coverage, and runtime efficiency over a total of 500 instances. 
All implementations of the approaches will be shared publicly.
%


\subsection{Performance of the MIP-framework}
\begin{figure*}[h]
  \centering
    \includegraphics[width=\textwidth]{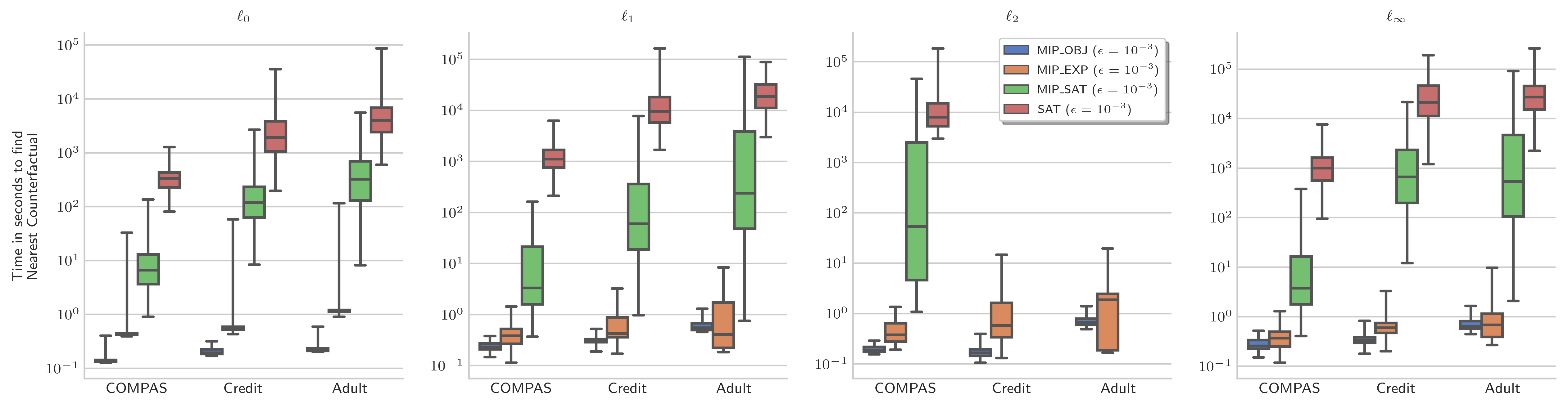}
  \caption{Full-setting runtime comparison of two-layer ReLU-activated NN with 10 neurons in each layer among our approach and MACE (SAT) \cite{mace}. Note that coverage is perfect by design. Each setting has been evaluated on 500 instances, however, SAT and MIP-SAT timed out on some samples. For such cases, only the samples for which all approaches have successfully finished running are included.}
  \label{fig:full-settings-time}
\end{figure*}

In the first set of experiments, we aim to showcase the ability of the proposed MIP-based approaches (i.e., MIP-SAT, MIP-EXP, MIP-OBJ) in diverse settings.
Specifically, we generate CFEs for a two-layer ReLU-activated NN with 10 neurons in each layer and evaluate generated counterfactual explanations using the metrics above on three datasets and four norm distances: $\ell_0, \ell_1, \ell_2, \ell_\infty$.
As expected, the CFE distances for all presented methods are similar to those of  SAT~\cite{mace}, which we use here as oracle, and coverage is perfect by design for all presented methods.
Figure~\ref{fig:full-settings-time} presents a comparison of runtime for these methods, where we observe significant improvement in runtime compared to SAT-oracle. Similar comparison for distances may be found in Figure~\ref{fig:full-settings-distance} in the Appendix.
Importantly, the presented MIP-based methods are able to generate CFEs in settings in which neither MACE (SAT) nor MIP-SAT are able (e.g., Adult or Credit dataset on $\ell_2$ norm).

\begin{figure}[t]
\centering
    \includegraphics[width=0.49\linewidth]{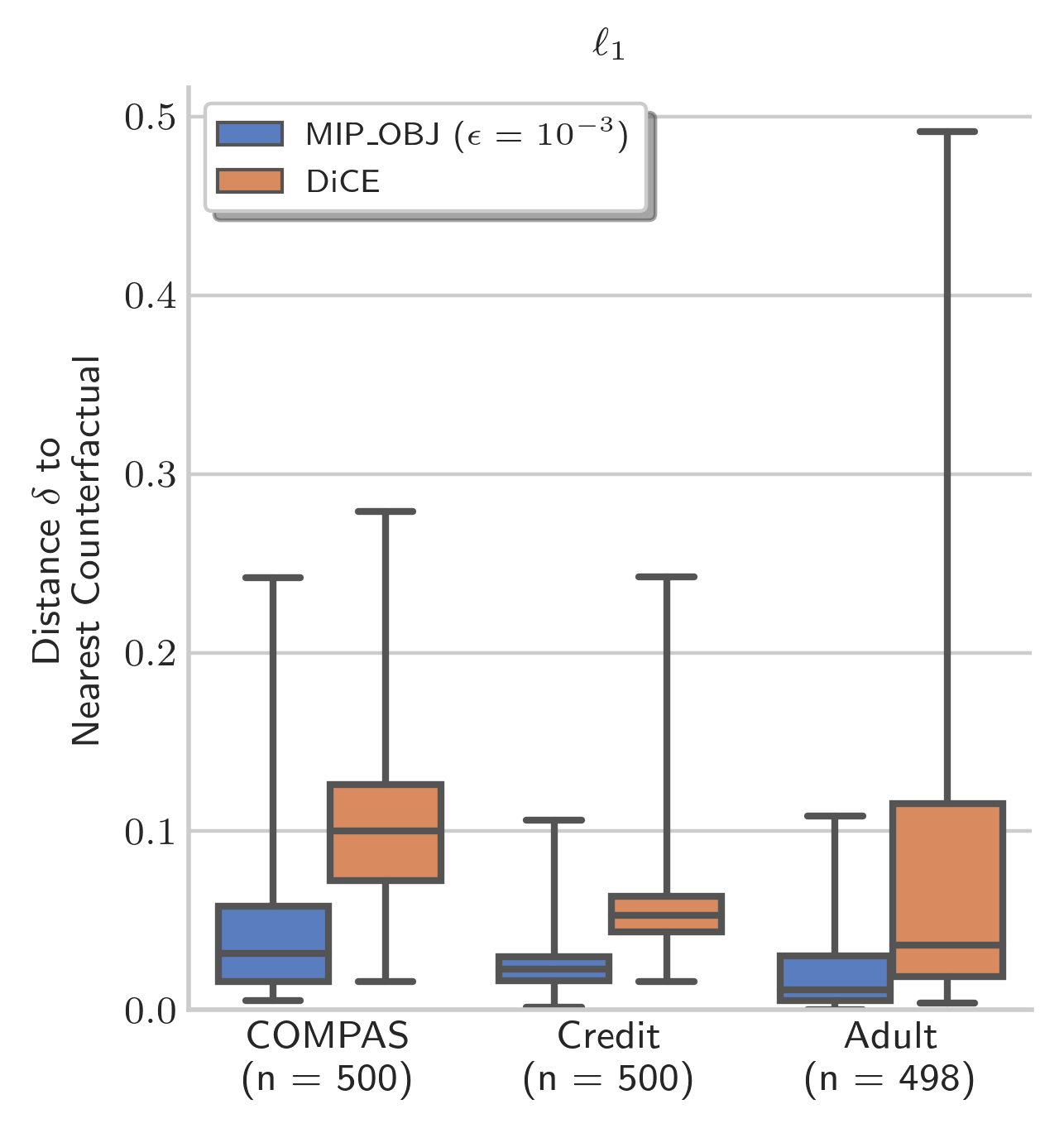}
    \includegraphics[width=0.49\linewidth]{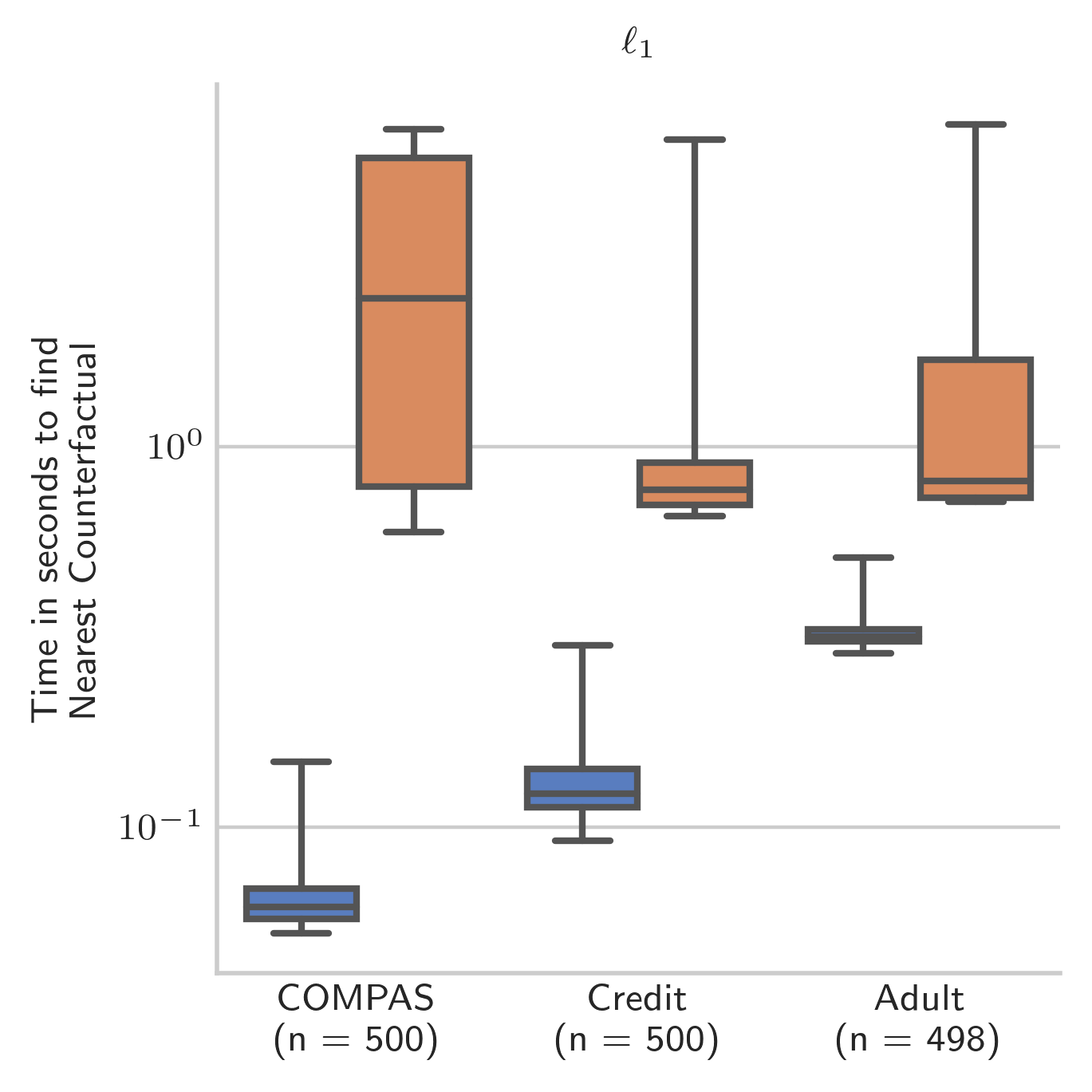}
  \caption{Distance and time comparison against DiCE as a gradient-based optimization approach. The model is a two-layered ReLU-activated NN with 10 neurons in each layer. MIP-OBJ coverage is perfect by design and DiCE coverage is also perfect except for Adult dataset (99.6\%).}
  \label{fig:mipobj_dice}
\end{figure}

In a second experiment, we compare the proposed MIP-based approaches, not only with the SAT-oracle but also with DiCE~\cite{dice} (i.e., gradient-based optimization) on the same NN model as above. \footref{fn:dice}
Here we adapt our experimental setting to  DiCE, as it only supports the $\ell_1$-norm distance, and does not provide support for ordinal and real-valued features.
Moreover, since DiCE assumes that the model has been trained using range-normalized data, we build additional support in our implementation to encode the normalization term in the MIP-based approaches, which in turn could negatively affect runtime and numeric stability.
Nonetheless, in this setting, we observe in Figure~\ref{fig:mipobj_dice} relatively smaller distances and significantly smaller runtimes for the former.
Furthermore, where MIP-OBJ has perfect coverage by design, DiCE dips slightly below perfect coverage on the Adult dataset, failing to offer an explanation for 2/500 instances.

\subsection{Scalablity Experiments}

The experiments above were presented on NN models that were able to sufficiently discriminate between the classes of the supervised learning task (with test accuracy in the range of 67-82\% for different datasets).
Complementing the demonstrations above, we investigate the scalibility of our approaches for the sake of completeness.
In this regard, Figure~\ref{fig:scalibility} (and Figure~\ref{fig:scalability_appendix} in the Appendix) compare the runtime, distance, and coverage for SMT-based~\cite{mace} and gradient-based~\cite{dice} approaches with our proposed approaches for a NN model with growing width and/or depth (as well as growing input size by incorporating different datasets).

It can be seen that the SMT-based approaches quickly reach their limit while MIP-based and gradient-based approaches scale well with both increasing width and depth. As MIP-based approaches do not scale polynomially w.r.t. network size, they do not scale as well as the gradient-based DiCE (this can be seen for the bigger Credit and Adult datasets in Figure \ref{fig:scalability_appendix} in the Appendix), however, they produce much smaller distances. 
While  MIP-based approaches have perfect coverage and minimum distance in theory, in practice numerical instabilities may be incurred in the backend tool as the number of intermediate variables in the mixed-integer program becomes large and their relations become deep due to the nested nature of NNs (the analysis of such numerical instabilities is beyond the scope of this work and deferred for future work). This causes failure to generate explanations for some samples or an increase in distances. In this context, having two MIP-based approaches is beneficial to verify results--for example, MIP-EXP behaves more stable in terms of distances than MIP-OBJ. 

\begin{figure*}[h]
\centering

    \includegraphics[width=0.8\linewidth]{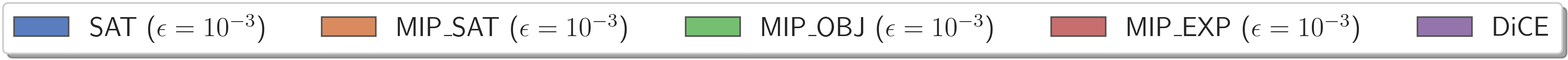}
    \includegraphics[width=0.35\linewidth]{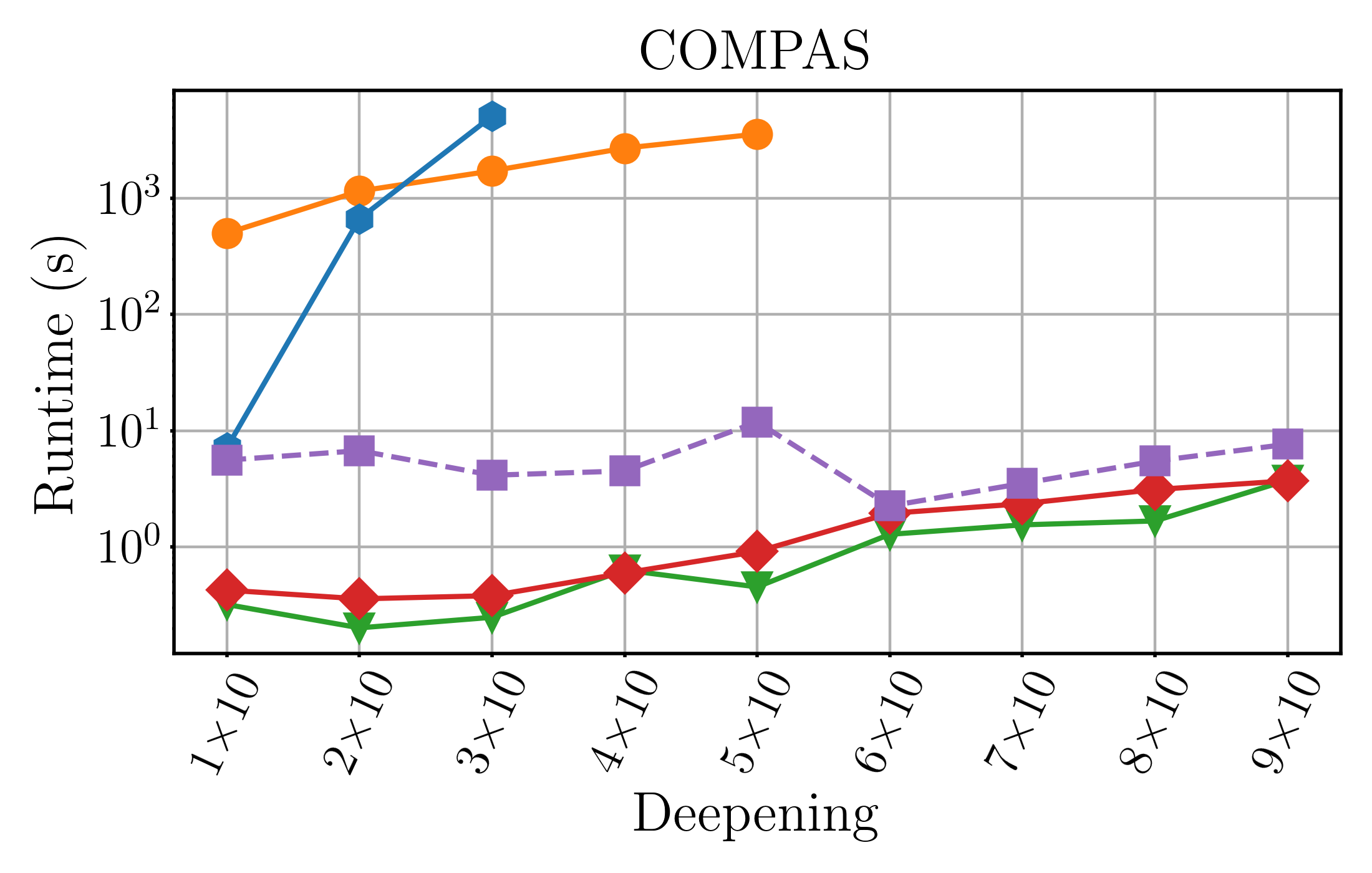}
    \includegraphics[width=0.55\linewidth, height=4cm]{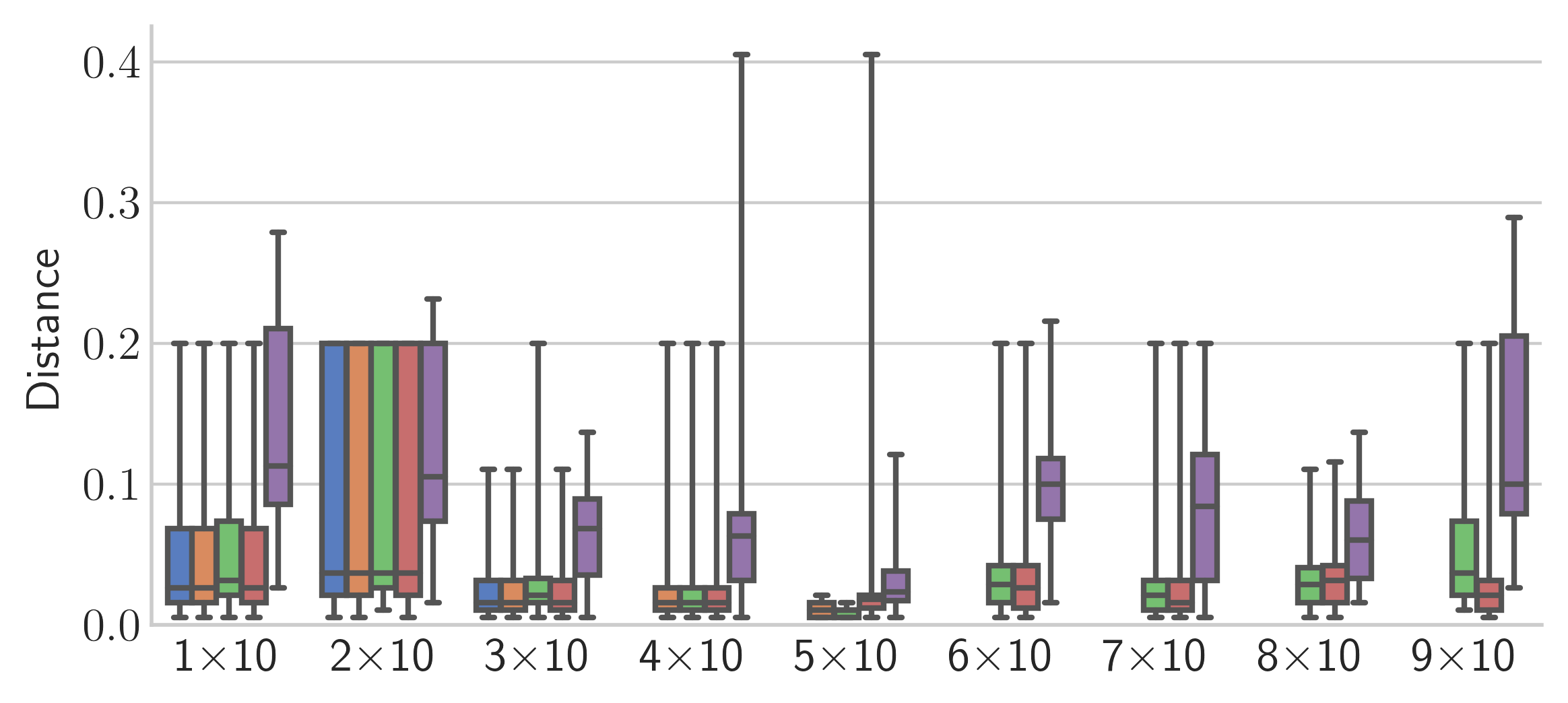}
    \includegraphics[width=0.35\linewidth]{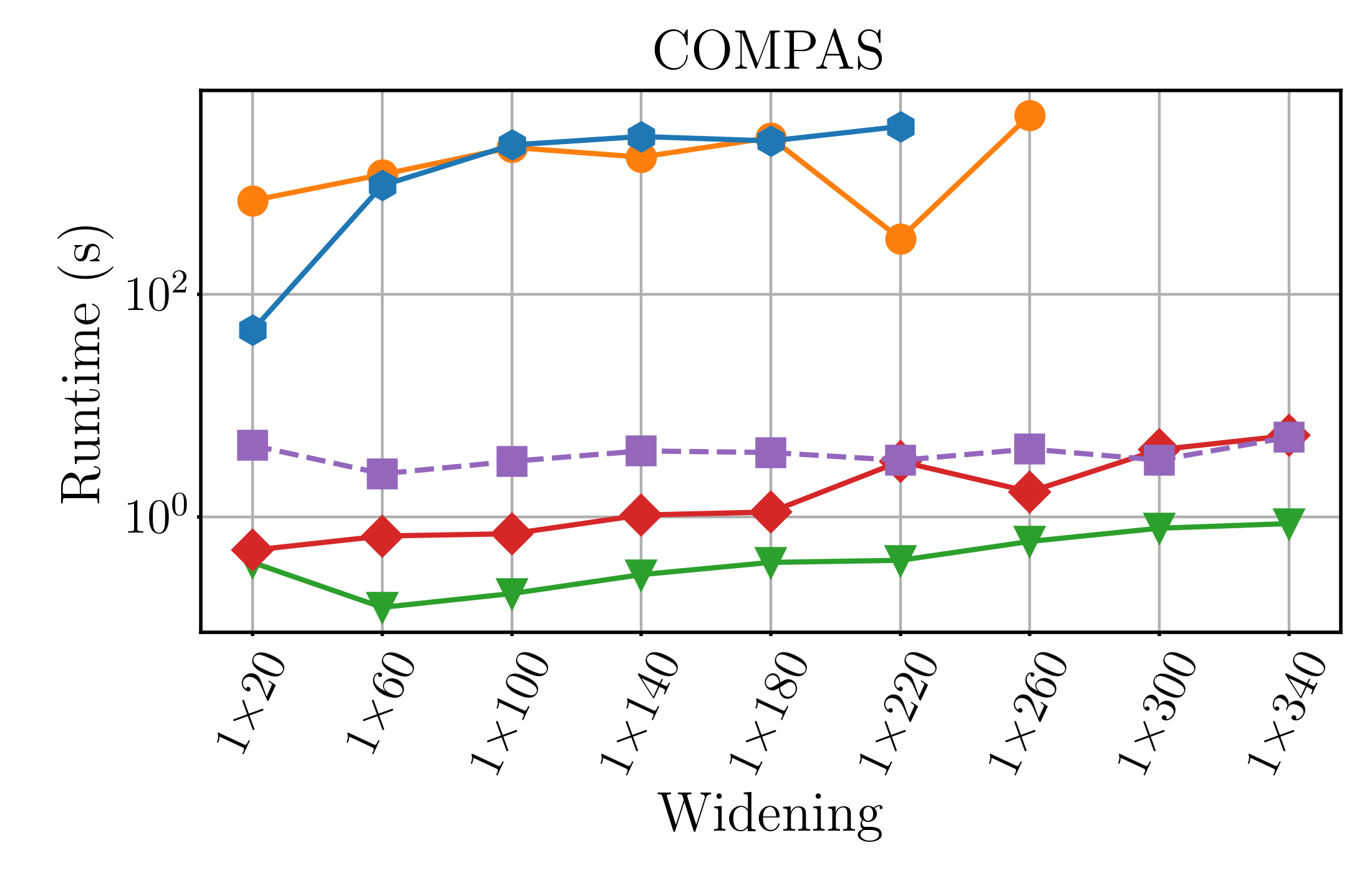}
    \includegraphics[width=0.55\linewidth, height=4cm]{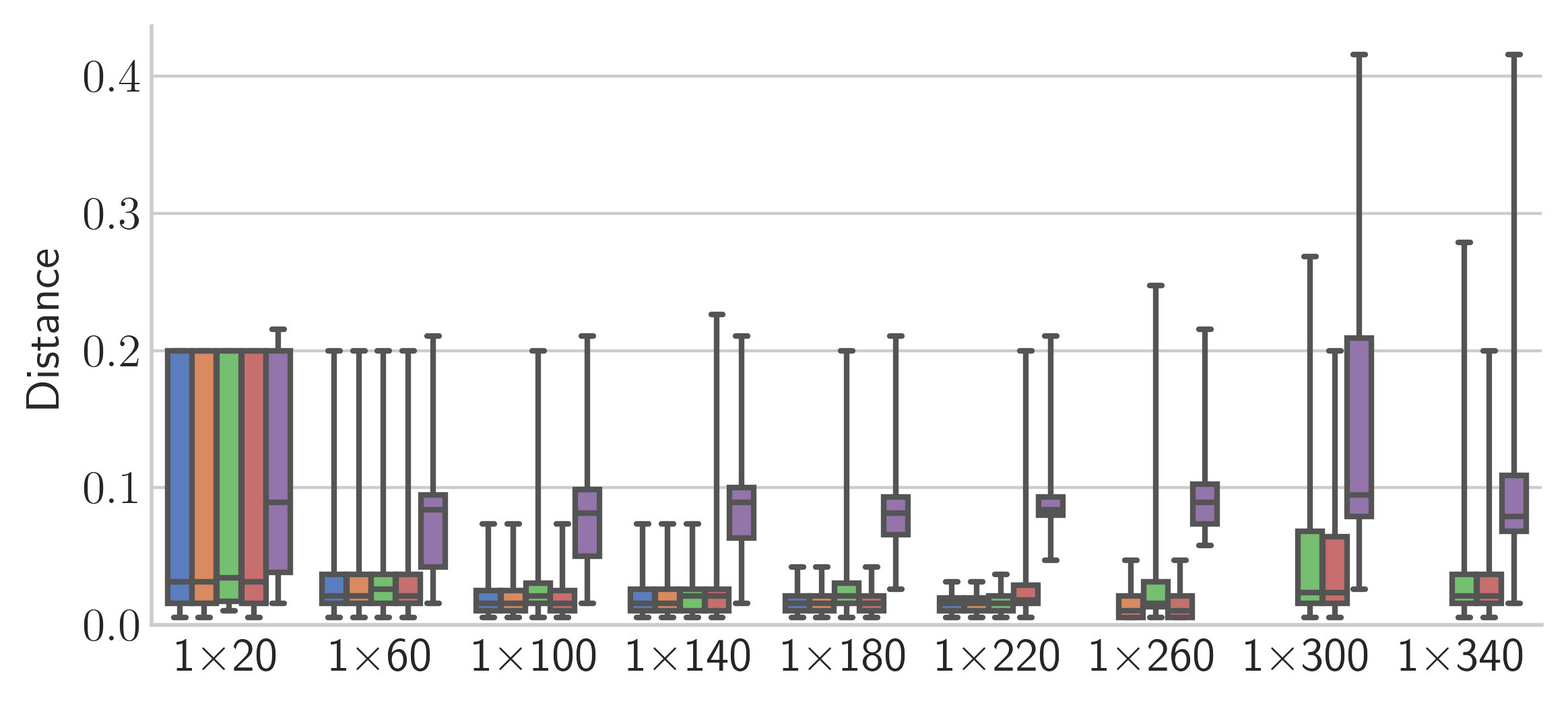}

  \caption{Scalability experiments comparing SMT-, MIP-, and gradient-based approaches on the COMPAS dataset. The upper row shows the results for increasing depth and the lower row for increasing width; both in terms of runtime and distance. For each approach and architecture 50 samples are evaluated, however, some fail to produce valid CFEs either because of imperfect coverage (i.e., DiCE) or numeric instabilities (i.e., MIP-OBJ and MIP-EXP); thus, only the instances for which all approaches have generated valid CFEs are included in the comparison. In general, for increasing depth, the average coverage across all the architectures is 99.1\% and 93.7\% for MIP-OBJ and MIP-EXP, and 96.4\% for DiCE. For increasing width, the average coverage across all the architectures is 100\% and 100\% for MIP-OBJ and MIP-EXP, and 100\% for DiCE. Similar experiments on the Credit and Adult datasets may be found in Figure \ref{fig:scalability_appendix} in the Appendix.}
  
  
  \label{fig:scalibility}
  
\end{figure*}


\subsection{Qualitative Experiments}

In this section, we show that how the expressiveness of SMT and MIP can be used to easily encode qualitative features and/or user-defined constraints for the explanations.

\subsubsection{\textbf{Diversity.}}
\label{sec:exp_diversity}
We report on experiments showing the diversity feature of our approach as presented in the previous section, and compare against DiCE's implementation of diversity.

We follow the authors of DiCE, and evaluate the $k$ diversely generated CFEs by measuring the mean of pairwise distances among the CFEs (the higher the better):

\begin{equation}
\label{eq:mean_diversity}
    k\!-\!diversity(\{\bold{x}_{j}^{CFE}\}_k): \frac{1}{{k \choose 2}} \sum_{i=1}^{k-1} \sum_{j=i+1}^{k} dist(\bold{x}_{i}^{CFE}, \bold{x}_{j}^{CFE})
\end{equation}

Expectedly, diversity is traded-off with distance. Thus, in addition to the diversity metric above, the distance of the diverse set of CFEs to the original factual instance, $\bold{x}^{F}$, is measured as follows (the lower the better):

\begin{equation}
\label{eq:mean_proximity}
    k\!-\!distance(\bold{x}^{F}, \{\bold{x}_{j}^{CFE}\}_k): \frac{1}{k} \sum_{i=1}^{k} dist(\bold{x}^{F}, \bold{x}_{i}^{CFE})
\end{equation}

\begin{figure*}[h!]
\centering
\includegraphics[width=0.9\linewidth]{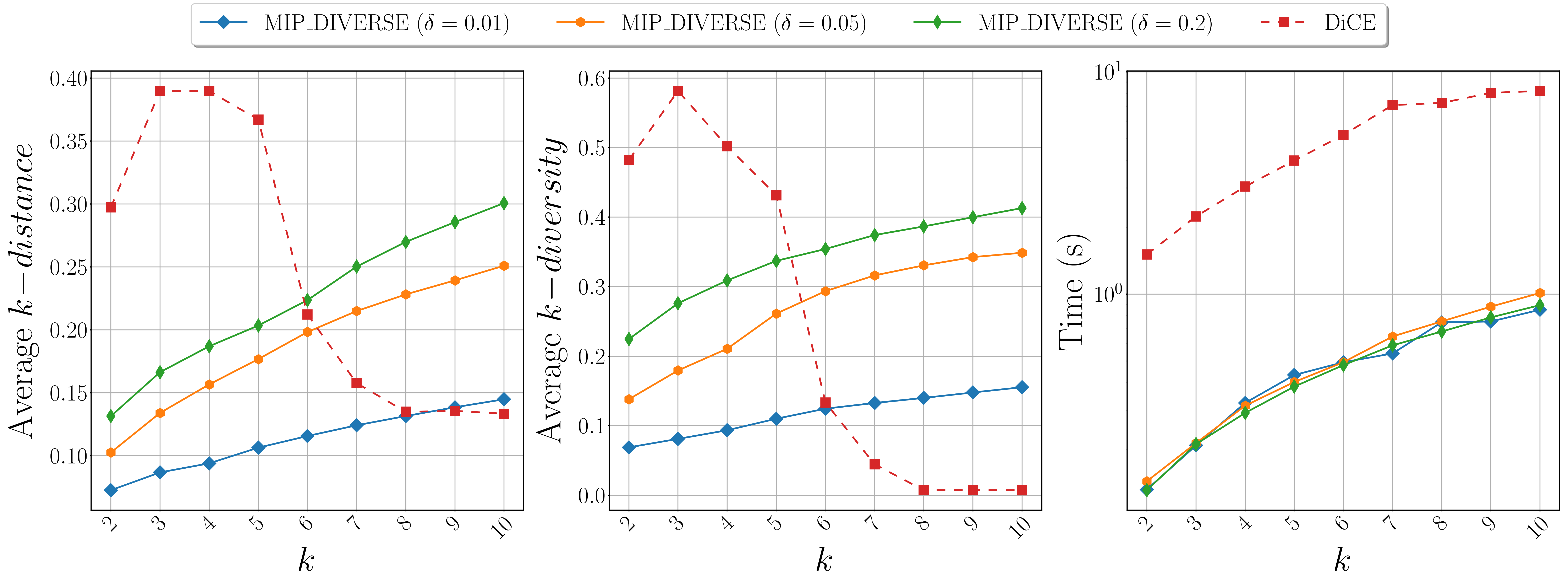}
\vspace{-10pt}
  \caption{Diversity, distance, and runtime for generating sets of counterfactuals on the COMPAS dataset and NN model with two hidden layers of size 10. For each counterfactual set size $k \in [2, 10]$, each approach has been tested on 100 instances. }
  \label{fig:diversity}
\end{figure*}

Figure \ref{fig:diversity} shows diversities generated by MIP-DIVERSE compared to DiCE for which the default hyperparameters are used. MIP-DIVERSE succeeds in finding the closest set of CFEs given a fixed distance threshold for diversity. The initial threshold has been set to 0.01 for this experiment, increasing it would result in the $k\!-\!diversity$ and $k\!-\!distance$ graph of Figure \ref{fig:diversity} to move upward, providing the possibility to choose the desired diversity-distance trade-off. Our results show that at a similar level of diversity (i.e., $k = 6$), the counterfactual set of MIP-DIVERSE is much closer to the factual instance. As $k$ increases further, in DiCE, while still a subset of the CFEs are diverse (and thus increase the average distance), the remaining ones are very similar to the previous as they minimally change a subset of the continuous variables. As a result, the average diversity and distance of the generated CFEs decreases. The runtimes of MIP-DIVERSE is again faster than the gradient-based opponent, however, MIP-DIVERSE is more sensitive to increasing the input size due to the added distance constraints, making it more or less as slow as DiCE on larger datasets.

\subsubsection{\textbf{Sparsity.}}
As described in the previous section, maximizing the sparsity of explanations is equivalent to minimizing the $\ell_0$ distance to the factual sample. To show the ability of our approach in maximizing sparsity, we refer the reader to the first column of figure \ref{fig:full-settings-distance} in the Appendix where all approaches succeed in maximizing sparsity. Indeed, it would also be possible to optimize for a convex combination of $\ell_0$ and e.g., $\ell_1$ norms to generate more realistic sparse explanations that allow more features to vary while staying close to the factual sample.

We would like to also remark, once more, the role of the expressive power of SMTs and MIPs, in increasing the quality of explanations through handling different types of constraints. For example, defining different types of actionability on the features (e.g., increase/decrease-only, non-actionable, etc.) are as simple as adding a few inequality constraints to the MIP model. This ease of encoding may give stake-holders and explanation-providers the possibility to take into account individual-specific situations where an individual might ask for her personal constraints to be considered within the provided explanation.

\section{Conclusion and Future Work}

In this work, we have proposed efficient approaches based on mixed-integer programming to generate counterfactual explanations with guarantees for the widely-used class of neural network models. We have empirically demonstrated the efficiency and guarantees of the proposed  framework by comparing it, in terms of distance, runtime and coverage with previous SMT- and gradient-based approaches for CFE generation. 
We have also provided qualitative results on the generation of diverse counterfactuals, showing the flexibility of our approach, as well as efficiency in handling complex qualitative features.

As future work, we plan to explore other qualitative features, such as other plausibility constraints beyond data types and ranges. 
Moreover, although in this work we have focused on NN architectures with ReLU activations, similar approaches can be deployed for any piece-wise linear activation function (e.g., Max-Pooling). Moreover, other classes of models (e.g., Support Vector Machines with RBF kernel)  could also be encoded or approximated by linear constraints, and thus be similarly  handled  by our MIP-framework.
Finally, as stake-holders increasingly adopt more complex neural models for consequential decision-making, it becomes critical to have access to reliable and efficient tools to explain algorithmic decisions. 
Thus, as venue for future work, it would be interesting to further investigate the scalability and numeric stability issues, which also arise in the NN verification.
\bibliography{biblio}

%
\appendix
\onecolumn


\section{Illustrations for the Bounds Computation} \label{sec:bounds_appendix}

\tikzset{
    mymx/.style={matrix of math nodes,nodes=myball,column sep=3.em,row sep=0ex},
    myball/.style={draw,circle,inner sep=4pt},
    mylabel/.style={near start,sloped,fill=white,inner sep=1pt,outer sep=1pt,below,
    execute at begin node={$\scriptstyle},execute at end node={$}},
    plain/.style={draw=none,fill=none},
    sel/.append style={fill=green!10},
    prevsel/.append style={fill=red!10},
    route/.style={-latex,thick},
    selroute/.style={route,blue!50!green}
}

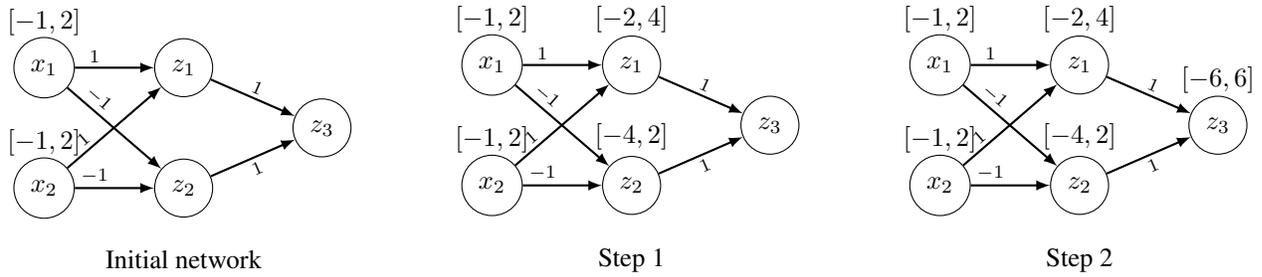
\begin{figure*}[h]
\hspace{-80pt}
\begin{minipage}[c]{0.33\linewidth}
\begin{center}
  \begin{tikzpicture}
    \matrix[mymx] (mx) {
      x_1        & z_1  \\
        &                  & z_3     \\
      x_2         & z_2  \\
    };
    {[route]
      \draw (mx-1-1) -- (mx-1-2) node[mylabel, above] { 1 };
      \draw (mx-1-1) -- (mx-3-2) node[mylabel, above] { -1 };
      \draw (mx-3-1) -- (mx-1-2) node[mylabel, above] { 1 };
      \draw (mx-3-1) -- (mx-3-2) node[mylabel, above] { -1 };
      \draw (mx-1-2) -- (mx-2-3) node[mylabel, midway, above] { 1 };
      \draw (mx-3-2) -- (mx-2-3) node[mylabel, midway, below] { 1 };
      
      \node [above=.3cm, align=flush center,text width=8cm] at (mx-1-1) {$[-1, 2]$};
      \node [above=.3cm, align=flush center,text width=8cm] at (mx-3-1) {$[-1, 2]$};
      
      \node [below=0.7cm, align=flush center,text width=8cm] at (mx-3-2) {Initial network};
    }
  \end{tikzpicture}

\end{center}
\end{minipage}
\begin{minipage}[c]{0.33\linewidth}
\begin{center}
  \begin{tikzpicture}
    \matrix[mymx] (mx) {
      x_1        & z_1  \\
        &                  & z_3     \\
      x_2         & z_2  \\
    };
    {[route]
      \draw (mx-1-1) -- (mx-1-2) node[mylabel, above] { 1 };
      \draw (mx-1-1) -- (mx-3-2) node[mylabel, above] { -1 };
      \draw (mx-3-1) -- (mx-1-2) node[mylabel, above] { 1 };
      \draw (mx-3-1) -- (mx-3-2) node[mylabel, above] { -1 };
      \draw (mx-1-2) -- (mx-2-3) node[mylabel, midway, above] { 1 };
      \draw (mx-3-2) -- (mx-2-3) node[mylabel, midway, below] { 1 };
      
      \node [above=.3cm, align=flush center,text width=8cm] at (mx-1-1) {$[-1, 2]$};
      \node [above=.3cm, align=flush center,text width=8cm] at (mx-3-1) {$[-1, 2]$};
      
      \node [above=.3cm, align=flush center,text width=8cm] at (mx-1-2) {$[-2, 4]$};
      \node [above=.35cm, align=flush center,text width=8cm] at (mx-3-2) {$[-4, 2]$};
      
      \node [below=0.7cm, align=flush center,text width=8cm] at (mx-3-2) {Step 1};
    }
  \end{tikzpicture}

\end{center}
\end{minipage}
\begin{minipage}[c]{0.33\linewidth}
\begin{center}
  \begin{tikzpicture}
    \matrix[mymx] (mx) {
      x_1        & z_1  \\
        &                  & z_3     \\
      x_2         & z_2  \\
    };
    {[route]
      \draw (mx-1-1) -- (mx-1-2) node[mylabel, above] { 1 };
      \draw (mx-1-1) -- (mx-3-2) node[mylabel, above] { -1 };
      \draw (mx-3-1) -- (mx-1-2) node[mylabel, above] { 1 };
      \draw (mx-3-1) -- (mx-3-2) node[mylabel, above] { -1 };
      \draw (mx-1-2) -- (mx-2-3) node[mylabel, midway, above] { 1 };
      \draw (mx-3-2) -- (mx-2-3) node[mylabel, midway, below] { 1 };
      
      \node [above=.3cm, align=flush center,text width=8cm] at (mx-1-1) {$[-1, 2]$};
      \node [above=.3cm, align=flush center,text width=8cm] at (mx-3-1) {$[-1, 2]$};
      
      \node [above=.3cm, align=flush center,text width=8cm] at (mx-1-2) {$[-2, 4]$};
      \node [above=.35cm, align=flush center,text width=8cm] at (mx-3-2) {$[-4, 2]$};
      
      \node [above=.3cm, align=flush center,text width=8cm] at (mx-2-3) {$[-6, 6]$};
      
      \node [below=0.7cm, align=flush center,text width=8cm] at (mx-3-2) {Step 2};
    }
  \end{tikzpicture}

\end{center}
\end{minipage}

\caption{Computing bounds using interval arithmetic}
\label{fig:bounds_computation}

\end{figure*}

We use a very simple example to demonstrate how bounds of the hidden units are computed using interval arithmetic and why using MIPs we can obtain better bounds. Consider the simple initial network without ReLUs and biases in Figure~\ref{fig:bounds_computation}. In step 1, we wish to compute the bounds for the first (and only) hidden layer. Starting by $z_1$, computing its lower bound means choosing the bounds from neurons of the previous layer which result in the minimum value for $z_1$. Thus, considering the sign of its weights, for both of the neurons in the previous layer the lower bound is chosen and the lower bound of $z_1$ is set to $1*(-1) + 1*(-1) = -2$. Similarly, the upper bound is $1*2 + 1*2 = 4$. For $z_2$, however, since the weights connected to it are negative, for computing lower bound, the upper bounds of previous layer are chosen and its lower bound is set to $-1*2 + -1*2 = -4$. Similarly, the upper bound is $-1*(-1) + -1*(-1) = 2$. Finally, in step 2, the bounds of the single output is computed in a similar way ($[-6, 6]$).

It can be seen that, in order to compute the bounds of the hidden layer, each neuron has chosen lower/upper bounds from the previous layer separately and without considering the relations among neurons, causing conflicts which result in loose bounds for the next layer (the output). On the other hand, considering the straight-forward MIP for this network, we simply have $z_1 = x_1 + x_2$ and $z_2 = -x_1 - x_2$ for the hidden layer and $z_3 = z_1 + z_2$ for the next layer. maximizing/minimizing $z_1$ and $z_2$ variables gives the same bounds as the ones by interval arithmetic for the hidden layer, however, for the next layer (the output) we will have the bounds $[0, 0]$ since the deeper relations among neurons are considered in the MIP i.e., $z_3 = z_1 + z_2 = x_1 + x_2 - x_1 - x_2 = 0$.

This example was for a network without the ReLU activation. The ReLUs can also be encoded by associating them with binary variables in the MIP encoding (e.g., encoding \eqref{eq:mip_unbound}) and compute \emph{exact} bounds similarly by solving MIPs layer-by-layer. However, this would be inefficient as the ReLU binary variables incur an exhaustive search. Thus, a linear (over-)approximation for ReLUs \eqref{eq:approx_b} is suggested to find looser than exact but tighter than interval arithmetic bounds in an efficient way.
%


\section{Additional Experiments}

\begin{figure*}[h]
  \centering
    \includegraphics[width=\textwidth]{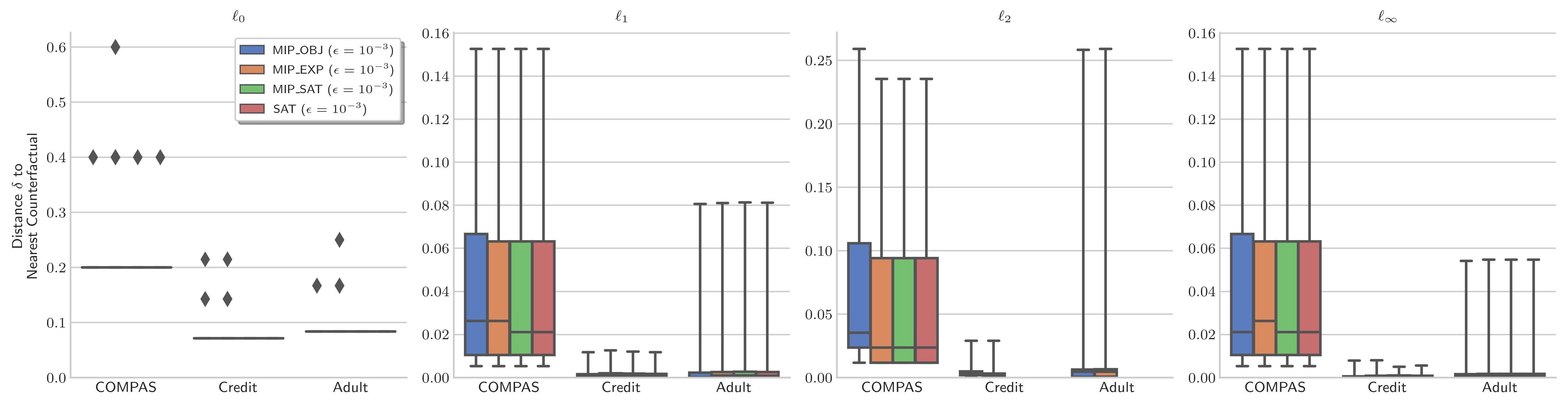}
  \caption{Full-setting distance comparison of two-layer ReLU-activated NN with 10 neurons in each layer among our approach and MACE (SAT) \cite{mace}. Note that coverage is perfect by design. Each setting has been evaluated on 500 instances, however, SAT and MIP-SAT timed out on some samples. For such cases, only the samples for which all approaches have successfully finished running are included.}
  \label{fig:full-settings-distance}
\end{figure*}

\begin{figure*}[t]
\centering
    \includegraphics[width=0.8\linewidth]{plots/fig4_legend.png}
    \includegraphics[width=0.35\linewidth]{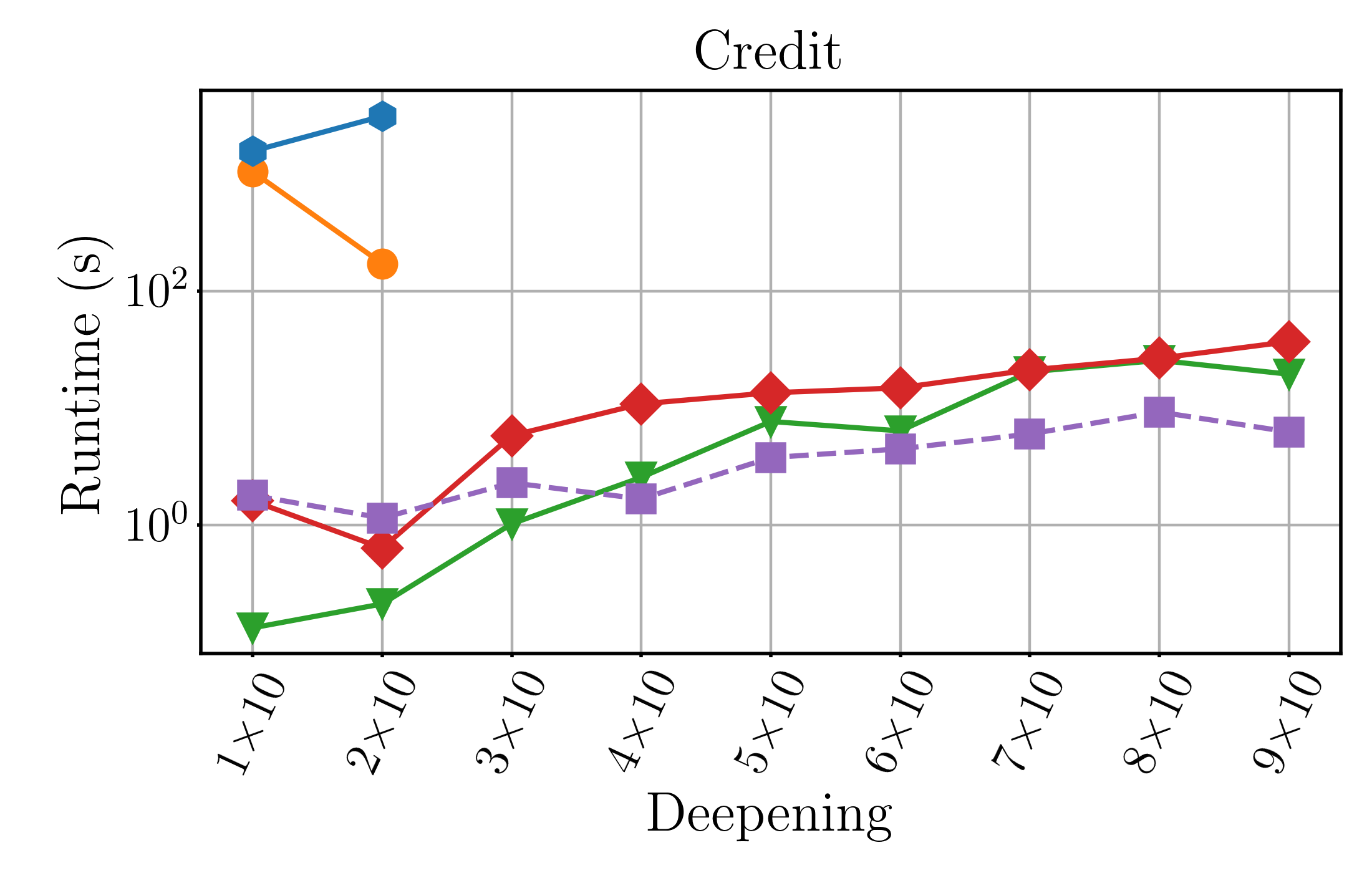}
    \includegraphics[width=0.55\linewidth]{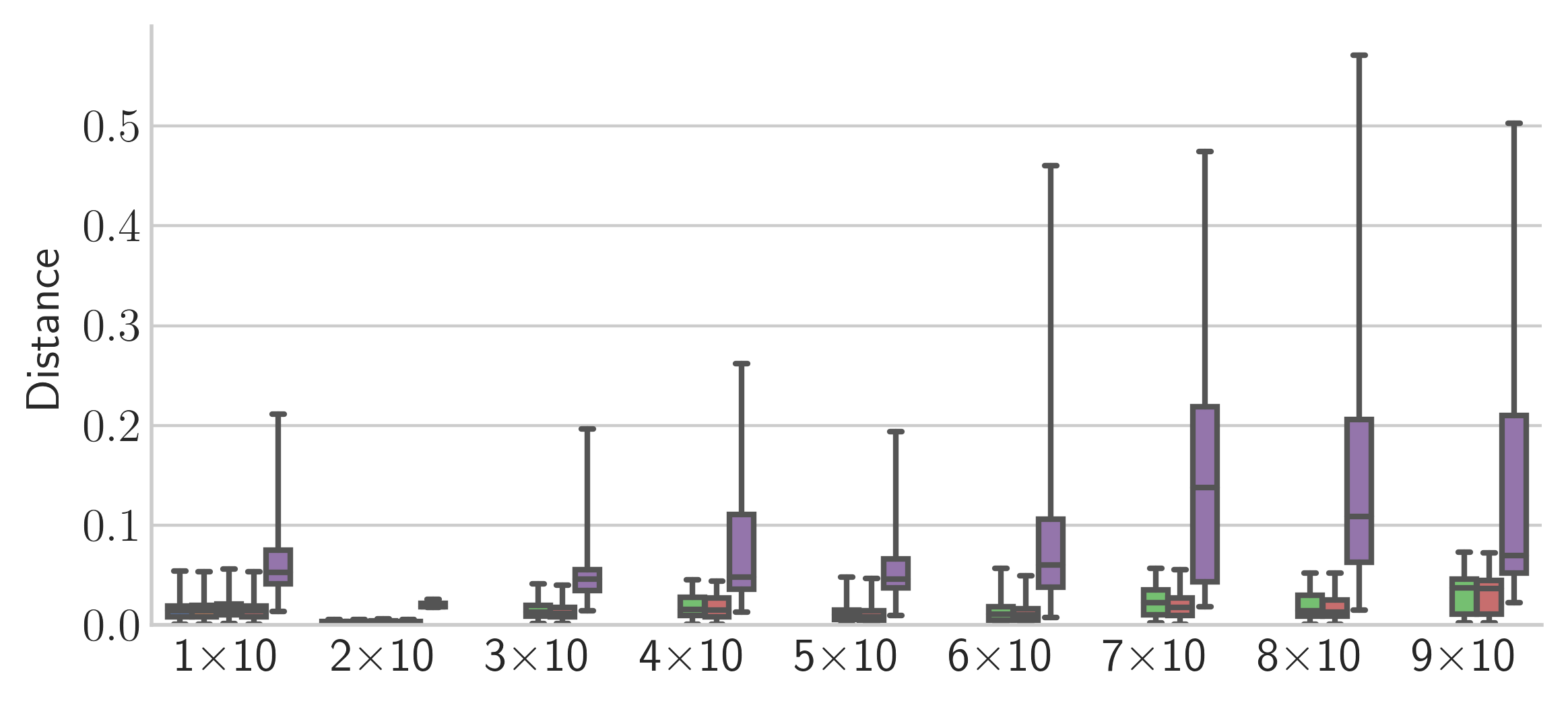}
    \includegraphics[width=0.35\linewidth]{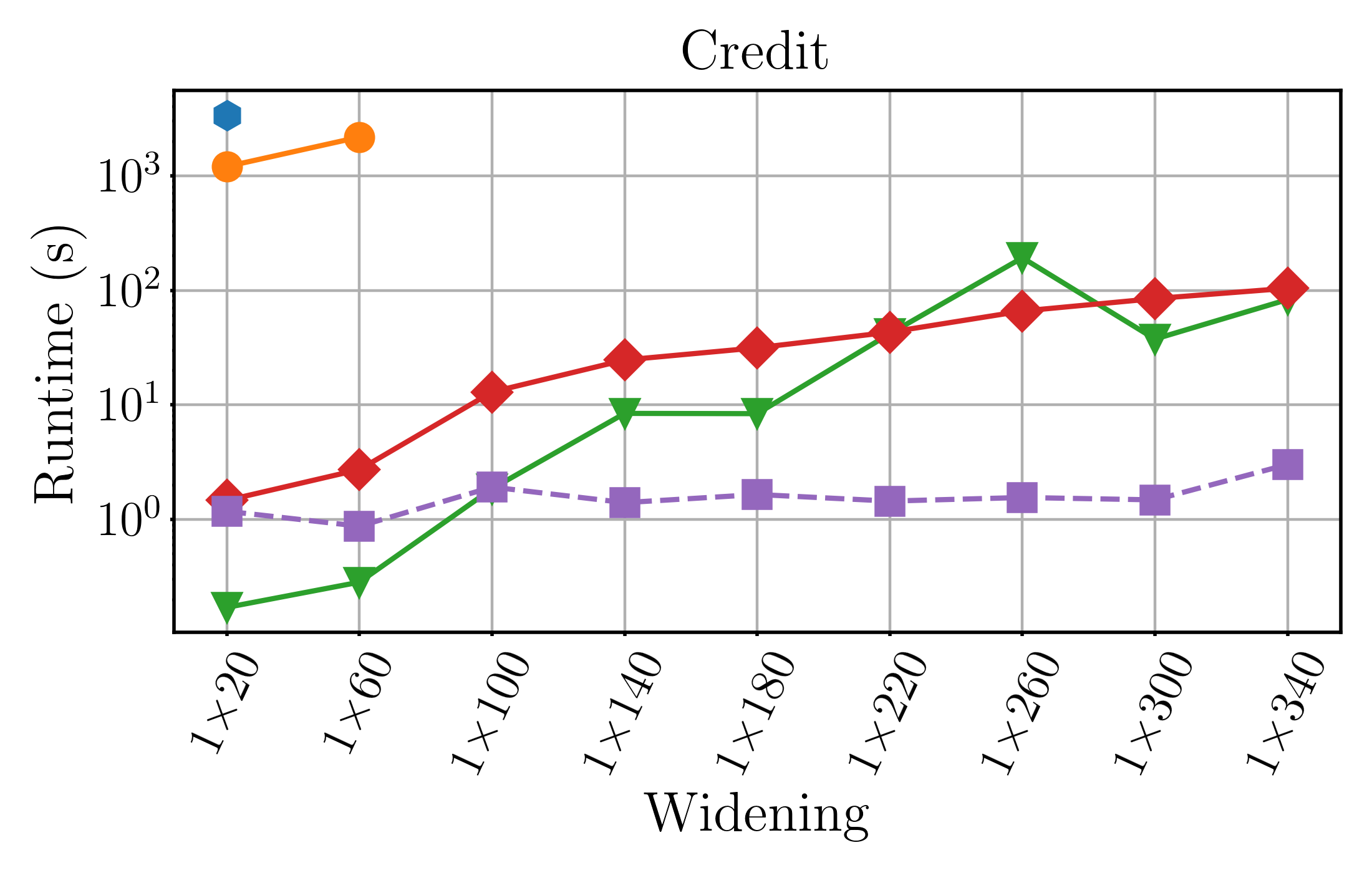}
    \includegraphics[width=0.55\linewidth]{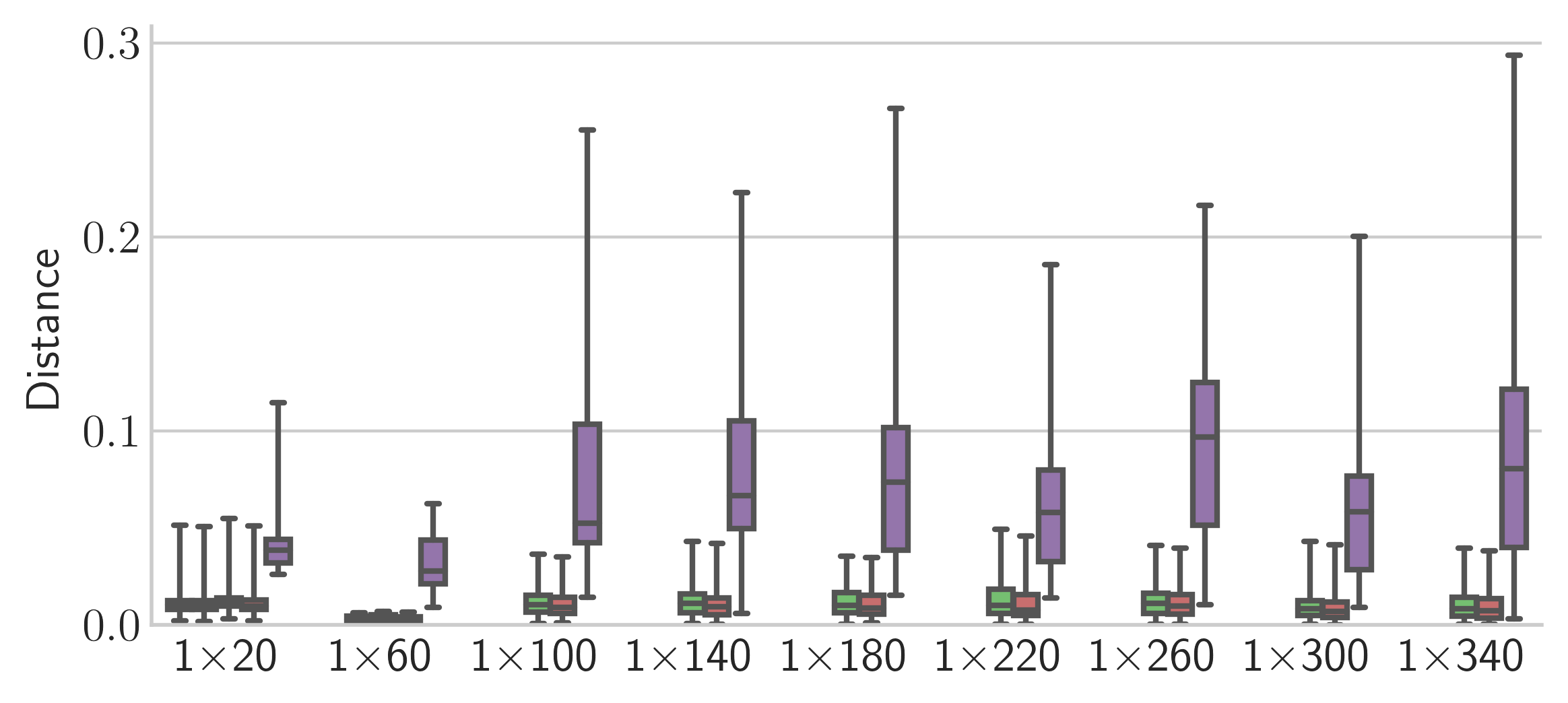}
    \includegraphics[width=0.35\linewidth]{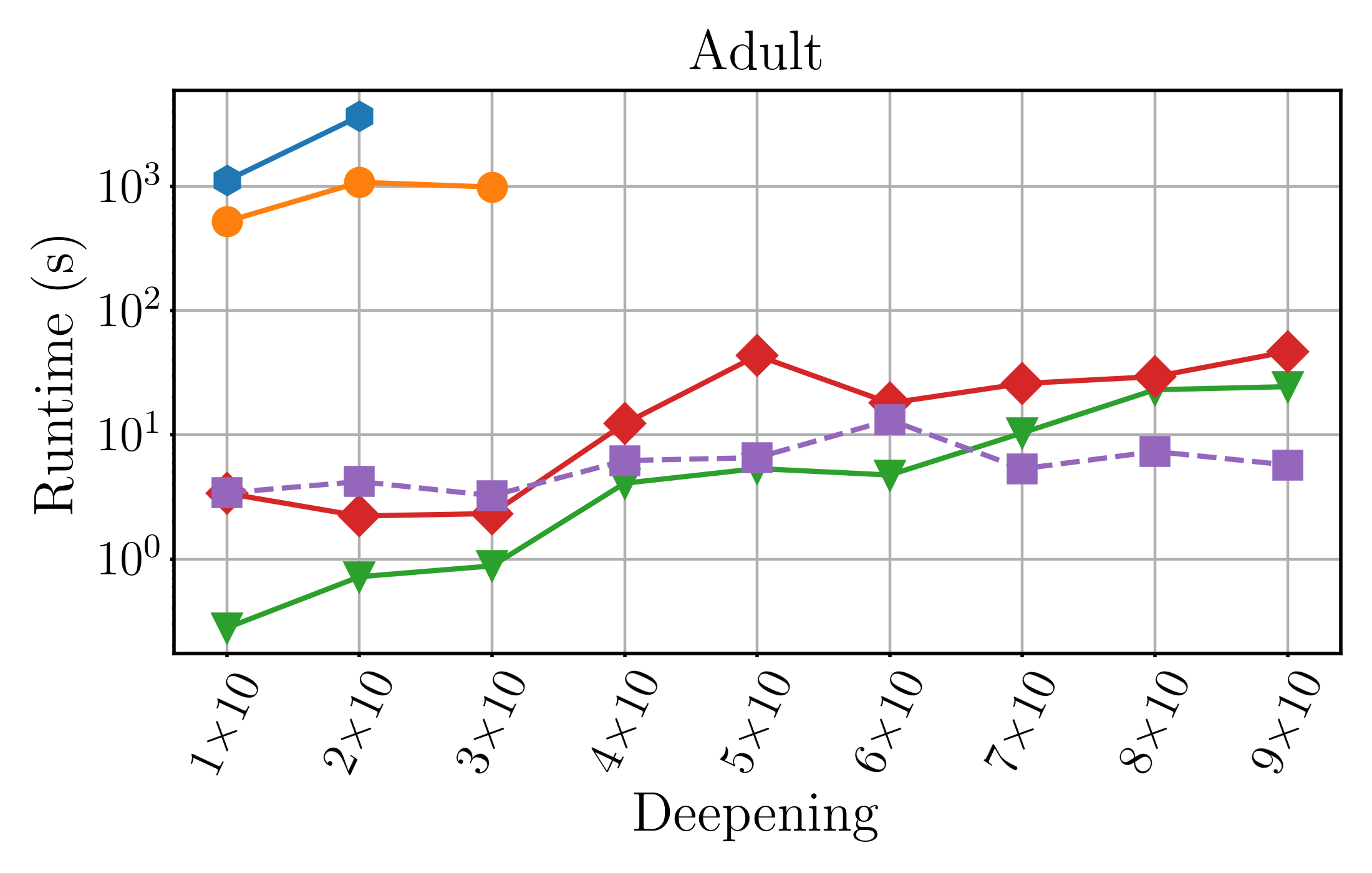}
    \includegraphics[width=0.55\linewidth]{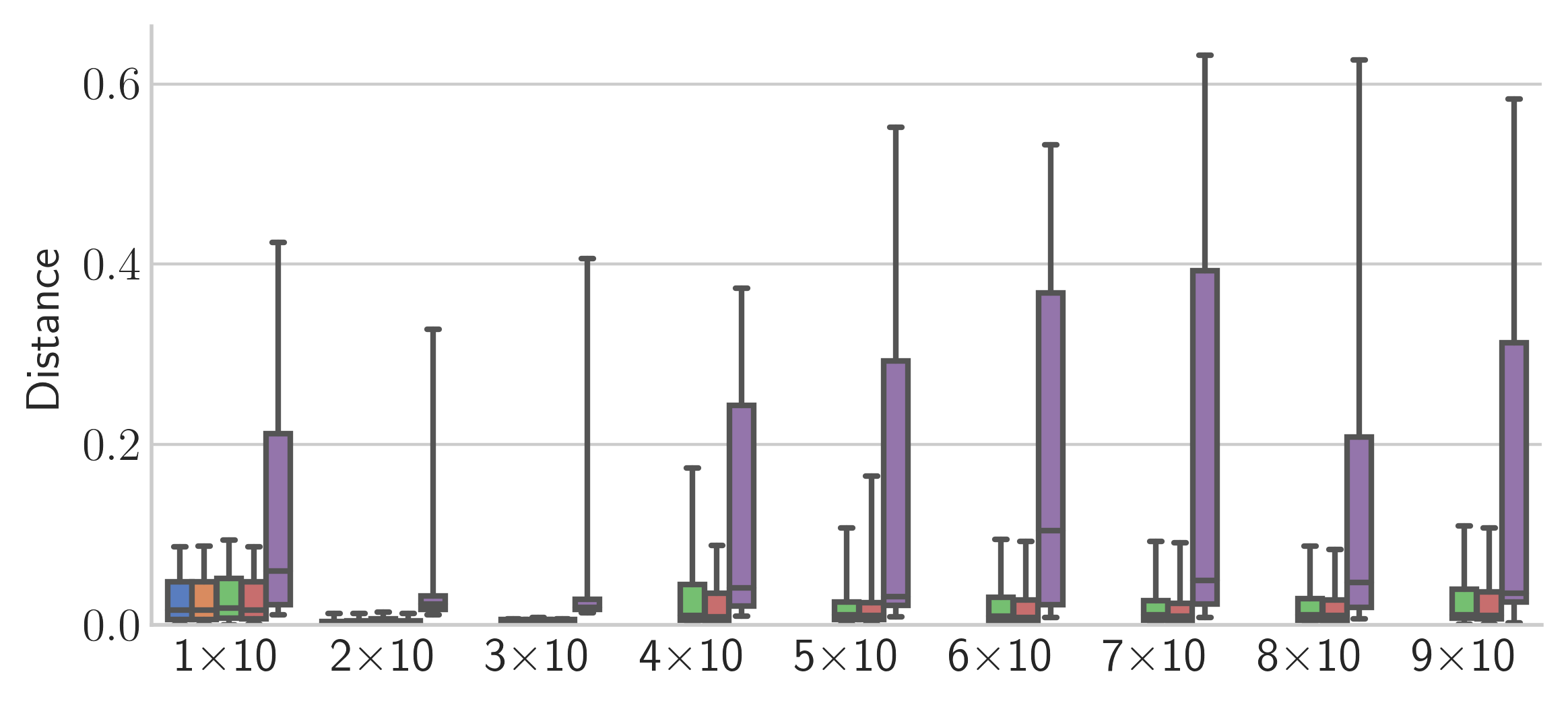}
    \includegraphics[width=0.35\linewidth]{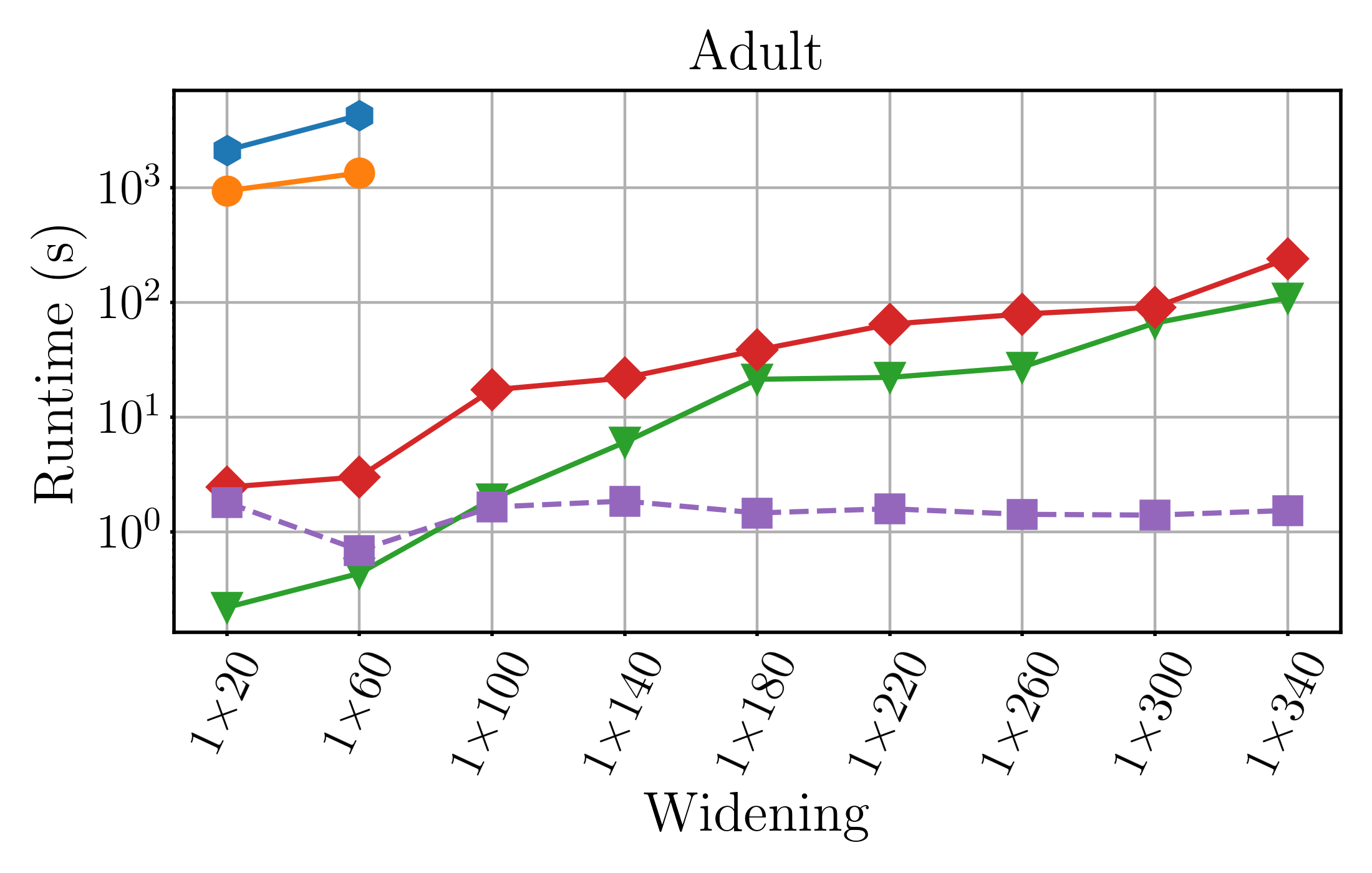}
    \includegraphics[width=0.55\linewidth]{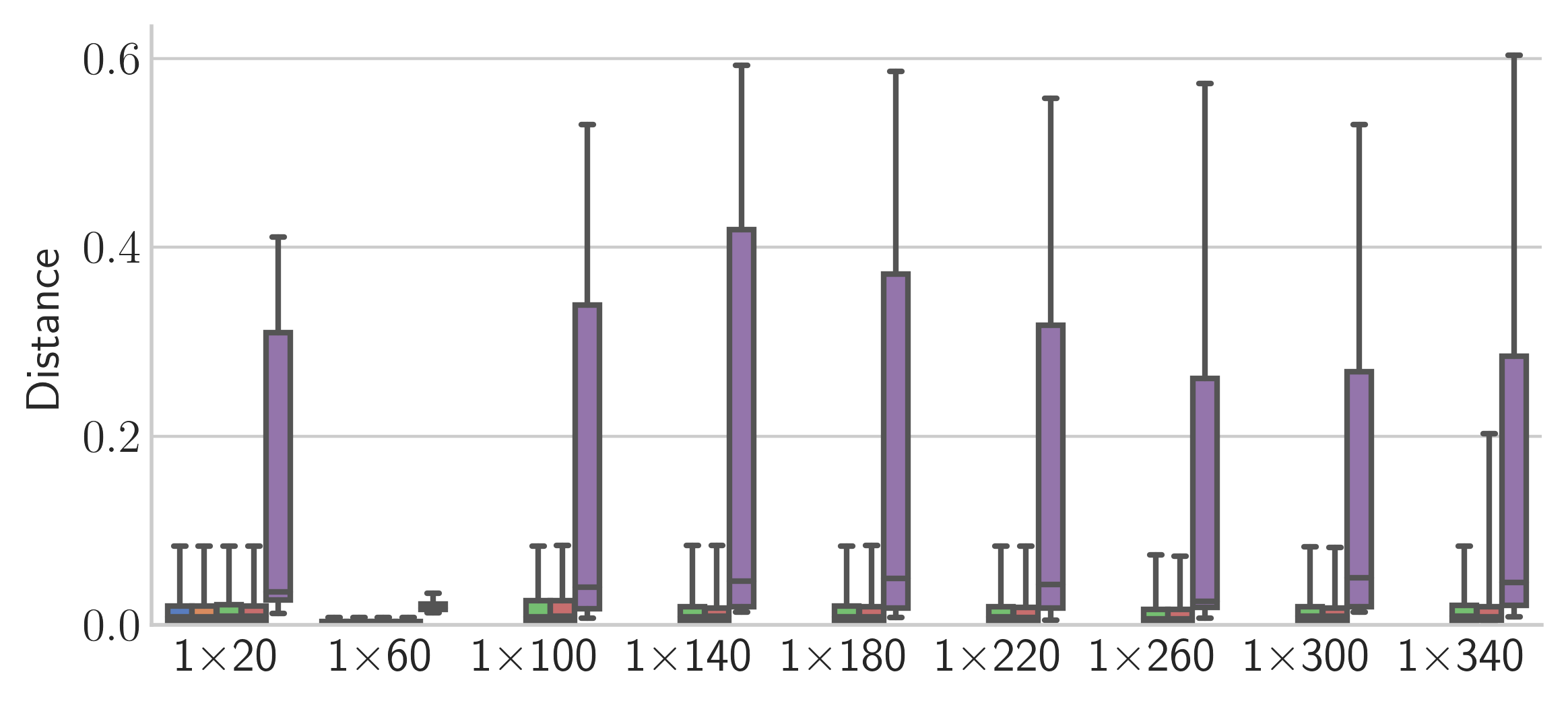}
  \caption{Scalability experiments comparing SMT-, MIP-, and gradient-based approaches. The first two rows show the results for Credit dataset and the second two rows are for the Adult dataset. In each two rows, the upper row demonstrates increasing depth while the lower row demonstrates increasing width; both in terms of runtime and distance. For each approach and architecture 50 samples are evaluated, however, some fail to produce valid CFEs (only for DiCE in this case); thus, only the instances for which all approaches have generated valid CFEs are included in the comparison. In general, for the Credit dataset, increasing depth results in 100.0\%, 100.0\%, and 98.2\% average coverage and increasing width results in 100\%, 100\%, and 100.0\% average coverage for MIP-OBJ, MIP-EXP, and DiCE, respectively. For the Adult dataset, increasing depth results in 100.0\%, 100.0\%, and 96.8\% average coverage and increasing width results in 100\%, 100\%, and 99.1\% average coverage for MIP-OBJ, MIP-EXP, and DiCE, respectively.}
  
  \label{fig:scalability_appendix}
\end{figure*}

The results in Figure~\ref{fig:full-settings-distance} complement those in Figure~\ref{fig:mipobj_dice} in the main body, by comparing instead the distance norm obtained by every method.
Additionally, Figure~\ref{fig:scalability_appendix} presents additional scalability results (similar to Figure~\ref{fig:scalibility}) but for the Adult and Credit datasets.
These results mimic the same trends seen earlier in the main body.

\end{document}